\documentclass[lettersize,journal]{IEEEtran}

\usepackage{amsmath}
\usepackage{epsfig}
\usepackage{multirow}
\usepackage{color}
\usepackage[ruled,linesnumbered,noend]{algorithm2e}
\usepackage{algpseudocode} 
\renewcommand{\algorithmiccomment}[1]{\bgroup\hfill\footnotesize$\triangleright$~#1\egroup}
\usepackage{placeins}

\usepackage{cite}
\usepackage{bm}
\usepackage{amsfonts}
\usepackage{graphics}
\usepackage{url}
\usepackage{soul}
\usepackage{mathrsfs}
\usepackage{flushend}
\usepackage{float}
\usepackage{tabularx}
\usepackage{array}
\usepackage{ragged2e}
\newcolumntype{P}[1]{>{\centering\hspace{0pt}}p{#1}} 
\newcolumntype{Z}{>{\centering\let\newline\\\arraybackslash\hspace{0pt}}X} 

\usepackage{lipsum} 

\soulregister{\cite}7
\soulregister{\citep}7
\soulregister{\citet}7
\soulregister{\ref}7
\soulregister{\pageref}7
\soulregister{\uppercase\expandafter}7
\sethlcolor{yellow}

\begin{document}

\title{Continuous Occupancy Mapping in Dynamic Environments Using Particles}

\author{Gang~Chen, Wei~Dong, Peng~Peng, Javier~Alonso-Mora, and Xiangyang~Zhu
\thanks{Gang Chen, Wei Dong, Peng Peng, and Xiangyang Zhu are with
        the State Key Laboratory of Mechanical System and Vibration,
        School of Mechanical Engineering, Shanghai Jiaotong University,
        200240, Shanghai, China. E-mails: \{chg947089399, dr.dongwei, yc\_pengpeng, mexyzhu\}@sjtu.edu.cn.
        }
\thanks{Javier Alonso-Mora is with the Autonomous Multi-Robots Lab, Department of Cognitive Robotics,
        Delft University of Technology, 2628 CD, Delft, Netherlands. E-mail: j.alonsomora@tudelft.nl.
        }
\thanks{This work was supported in part by the National Natural Science Foundation of China Grant 51975348 and in part by Shanghai Rising-Star Program under Grant 22QA1404400.}
\thanks{Corresponding authors: Wei~Dong and Xiangyang Zhu.
        }}




\maketitle

\begin{abstract}
Particle-based dynamic occupancy maps were proposed in recent years to model the obstacles in dynamic environments.
Current particle-based maps describe the occupancy status in discrete grid form and suffer from the grid size problem, wherein a large grid size is unfavorable for motion planning while a small grid size lowers efficiency and causes gaps and inconsistencies.
To tackle this problem, this paper generalizes the particle-based map into continuous space and builds an efficient 3D egocentric local map.
A dual-structure subspace division paradigm, composed of a voxel subspace division and a novel pyramid-like subspace division, is proposed to propagate particles and update the map efficiently with the consideration of occlusions.
The occupancy status at an arbitrary point in the map space can then be estimated with the weights of the particles. 
To reduce the noise in modeling static and dynamic obstacles simultaneously, an initial velocity estimation approach and a mixture model are utilized.
Experimental results show that our map can effectively and efficiently model both dynamic obstacles and static obstacles.
Compared to the state-of-the-art grid-form particle-based map, our map enables continuous occupancy estimation and substantially improves the mapping performance at different resolutions.
\end{abstract}


\IEEEpeerreviewmaketitle

\section{Introduction} \label{Section: Introduction}

\IEEEPARstart{T}{he} particle-based map is originally proposed in \cite{SMCPHDMap} for dynamic and unstructured environments. Particles with position and velocity states are used to approximate both dynamic obstacles and static obstacles on the basis of sequential Monte Carlo (SMC) filtering. In recent works, \cite{RFSMap} introduces the theory of random finite set (RFS) to particle-based maps. The probability hypothesis density (PHD) filter is applied to predict and update the particles and estimate the dynamics of the grids in the map. Later, \cite{MixtureModelTUM,NewBenz,DynamicMapICRA2021} improve the particle-based maps by considering the mixture model, semantic information and high-level occupancy status inference, respectively.
Due to the ability to model complex-shaped static and dynamic obstacles simultaneously, particle-based maps draw more attention in representing dynamic environments. 
Currently, the input form of particle-based maps is the ray-casting-generated measurement grid map originated from the first work \cite{SMCPHDMap}, and thus the map is discretized with grids. 
This discrete form inhibits the state estimation resolution and brings the grid size problem, namely: large grids lead to a low resolution that is unfavorable for motion planning, while small grids increase the computation requirements and may cause gaps and inconsistencies \cite{ContinuousMapMIT}.
Besides, desktop GPUs are required to run the particle-based maps in real-time, and a more efficient map is needed for applications in small-scale robotic systems.

This work proposes a dual-structure particle-based (DSP) map, a continuous dynamic occupancy map free from the grid size problem.
The input of the map is the raw point cloud rather than the measurement grid map. A novel dual-structure map building paradigm, composed of a voxel subspace division for particle storage and resampling and a dynamic pyramid-like subspace division for occlusion-aware particle update, is proposed to model the local environment with particles that have continuous states. Under the Gaussian noise assumption, we demonstrate that this updating paradigm is effective and computationally efficient.
To reduce the noise in simultaneously modeling static and dynamic obstacles, the importance of newborn particles is addressed by using non-Gaussian initial velocity estimation and a mixture model that adaptively allocates the number of static and dynamic particles.
With a complete process of prediction, update, birth, and resampling of particles in the continuous space, the occupancy status at an arbitrary point in the map can be estimated using onboard CPU devices.

In the experimental tests, we first evaluated the dynamic obstacle velocity estimation precision of the map. Then the ablation study was conducted to identify the mapping parameters.
Subsequently, comparison tests were carried out, involving a state-of-the-art particle-based dynamic occupancy map \cite{DynamicMapICRA2021} and a widely used static occupancy map \cite{Ringbuffer}. Results show that our map has the best occupancy status estimation performance in dynamic environments and competitive performance with \cite{Ringbuffer} in static environments.
Furthermore, we verified the DSP map in obstacle avoidance tasks of a mini quadrotor in different environments.
To the best of the authors' knowledge, this is the first continuous particle-based occupancy map and the first dynamic occupancy map that can be applied to small-scale robotic systems like quadrotors.

The main contributions of this work include:
\begin{enumerate}
  \item A novel dual-structure particle-based map building paradigm that enables continuous mapping of the occupancy status in dynamic environments.
  \item The leverage of initial velocity estimation and an efficient mixture model to reduce noise in modeling static and dynamic obstacles simultaneously.
  \item The complete procedures of building a DSP map that can be applied to onboard computing devices of small-scale robotic systems.
  \item The released code at \url{https://github.com/g-ch/DSP-map}, including an example application in ROS.
\end{enumerate}

The remaining content is organized as follows: Section \ref{Section: Related work} describes the related work. Section \ref{Section: Preliminaries} presents the background knowledge of our map.
Section \ref{Section: world model} explains the formulations of the world model and gives an overview of mapping procedures. Section \ref{Section: Map Building} expresses the mapping procedures with the dual structure. In Section \ref{Section: extensions in mapping}, more components for mapping are discussed. Section \ref{Section: Implementation} presents some implementation details. The experimental results and the conclusion are described in Section \ref{Section: experiments} and Section \ref{Section: conclusion}, respectively.


\section{Related Work} \label{Section: Related work}
\subsection{Discrete Map and Continuous Map}
Environment representation is fundamental to obstacle avoidance of robotics systems. One of the most popular representation approaches is occupancy mapping, which originated from \cite{GridMapOldest} and is capable of modeling cluttered environments. Grid map (2D or 3D) is a kind of computationally efficient form to realize occupancy mapping. The environment is usually divided into discrete grids, and the occupancy status of each grid is updated with the ray casting algorithm \cite{Ringbuffer,OctoMap,UFOMap,FPGAOmni}. The size of the grids, however, is difficult to determine. Large grids lead to a low resolution that is unfavorable for motion planning. Small grids increase the computation requirements and cause gaps and inconsistencies when the input point clouds are sparse or noisy \cite{ContinuousMapMIT}. To avoid the grid size problem and allow arbitrary resolutions, the paradigm of building the map with continuous occupancy probability kernels rather than grids is proposed \cite{ContinuousMapMIT,GaussianProcessMap,StaticHilbert}. Free space and occupied points or segments are first generated with the input point clouds and then used to update the parameters in the kernel functions. The occupancy status at an arbitrary position can then be estimated with nearby kernels. 

\subsection{Occupancy Maps in Dynamic Environments} \label{Section: Occupancy Maps in Dynamic Environments}
The maps mentioned above  \cite{GridMapOldest,OctoMap,UFOMap,FPGAOmni,GaussianProcessMap,StaticHilbert} are built under the assumption that the environment is static. As the robotic systems were deployed in dynamic environments,  improvements have to be made to instantly represent the occupancy position of dynamic obstacles, such as pedestrians and other robots, and, even further, to predict the future positions of dynamic obstacles. An intuitive approach is to leverage independent detection and tracking of moving objects (DATMO) \cite{SLAMMOT, ActiveSensingOwn,DynamicObstalceAvoidanceTUD} to model the dynamic obstacles and utilize static occupancy maps still to represent the other objects. A prerequisite of DATMO is that the detection and shape models of the dynamic obstacles are well-trained \cite{MixtureModelTUM}, which conflicts with the unknown environment characters in many tasks. In addition, difficulties in data association \cite{MixtureModelTUM} and the trail noise caused by obstacles movements in the static map \cite{ActiveSensingOwn,NewIeeeAccess} are intractable. Therefore, improving the map itself directly by considering the dynamic obstacle assumption is required, and the dynamic occupancy map \cite{OldSimpleDynamicMap2,OldSimpleDynamicMap} emerges accordingly.


Early dynamic occupancy maps treat the dynamic obstacles, such as pedestrians and robots, as spurious data in the map, and detect and remove the data to build a robust static map \cite{OldSimpleDynamicMap2,OldSimpleDynamicMap,MarkovChainMap,NDTMap}. Starting from the latest decade  \cite{SMCPHDMap}, research works considering modeling the dynamics, mostly velocities, of the obstacles in the map have been carried out to improve the obstacle avoidance performance in dynamic environments. Various methods have been proposed in these works. Some apply the dynamic obstacle assumption to the existing structures of static occupancy maps. For example, \cite{DynamicGaussianMap} adopts optical-flow-based motion maps to estimate the velocity of grids and improves the Gaussian process occupancy map \cite{GaussianProcessMap} to adapt to dynamic environments. \cite{ImprovedDynamicGaussian} further improves \cite{DynamicGaussianMap} by learning dynamic areas with stochastic variational inference. In \cite{DynamicHilbert}, point clouds from lidar are clustered and filtered to estimate the velocities of dynamic obstacles. The estimation is applied to generate non-stationary kernels in the Hilbert space to build the dynamic Hilbert map. With the popularity of deep learning methods, some recent works adopt neural networks to predict the velocity of each grid in a grid map \cite{RNNDynamicMap0}, \cite{RNNDynamicMap} or future occupancy status \cite{NewIROS}, \cite{NewICRA}.

\subsection{Particle-based Dynamic Occupancy Maps}
The particle-based map originates from the autonomous driving area \cite{SMCPHDMap,2-5DSMCPHDMap}. In a particle-based map, an obstacle is regarded as a set of point objects and the particles with velocities are used to model the point objects. 
Compared to the dynamic occupancy maps in \ref{Section: Occupancy Maps in Dynamic Environments}, the particle-based map is originally proposed for dynamic environments and has a stronger potential to improve the mapping performance in complex and highly dynamic environments.
Nuss et al. \cite{RFSMap} improves \cite{SMCPHDMap,2-5DSMCPHDMap} by introducing the RFS theory and deriving map-building procedures with the PHD filter and the Bernoulli filter. The improved map can be built in real-time in 2D space with GPU devices. Later, \cite{DynamicMapICRA2021} generalizes \cite{RFSMap} to 3D space.

In a cluttered environment with dynamic and static obstacles, multiple point objects, dynamic or static, need to be modeled, and denoising is of great importance. Two approaches are usually adopted to reduce the noise. The first approach is to use a mixture model  \cite{DualPHDMap,MixtureModelDSTTUM,MixtureModelTUM,DynamicMapICRA2021}, which includes a separate static model and a dynamic model, to update the states of static and dynamic point objects independently. The mixture model works as dual PHD filters \cite{DualPHDMap} or the grid-level inference \cite{MixtureModelDSTTUM,MixtureModelTUM,DynamicMapICRA2021}. Another approach is to apply additional information to reduce the noise in the updating procedure. For instance, \cite{NewBenz} adds an extra semantic grid channel in the input to generate particles with semantic labels and update with the semantic association.

The above particle-based maps are still grid maps. Measurement grids generated by the ray casting method are adopted as the input, and the environment is described with discretized 2D or 3D girds. This discretized expression suffers from the grid size problem mentioned in Section \ref{Section: Introduction}. The grid size also limits the state estimation resolution of the obstacles. Therefore, a continuous particle-based occupancy map is required. In addition, since numerous particles are used, state-of-the-art particle-based maps usually rely on Desktop GPU devices for computation \cite{RFSMap,DynamicMapICRA2021}. To deploy the particle-based map on small-scale robotic systems, improving computational efficiency is necessary.

\begin{table}[h]
  \center
  \caption{Notation in this paper}
  \begin{tabularx}{3.5in}{|P{0.6in}|X|}
    \hline
    \textbf{Symbol} & \qquad  \qquad \qquad \qquad \quad \textbf{Meaning}                                        \\ \hline
  $\text{X}, \text{X}_k$      & RFS, RFS composed of point objects at time $k$. \\ \hline
  $\text{X}_k^{(\mathbb V_i)}, \text{X}_k^{(\mathbb{P}_i)}$      & RFS composed of point objects in a voxel subspace and a pyramid subspace, respectively, with index $i$ at time $k$. \\ \hline
  $\text{Z}_k$       & RFS composed of measurement points at time $k$. \\ \hline
  $\text{S}_{k|k-1}$       & RFS composed of survived objects from $k-1$ to $k$. \\ \hline
  $\text{B}_{k|k-1}$  & RFS composed of newborn objects from $k-1$ to $k$. \\ \hline
  $\text{O}_{k}, \text{C}_{k}$       & RFS composed of the detected objects and clutter at $k$. \\ \hline
  $\boldsymbol{x}^{(i)}$       & State vector of an element or an object with index $i$. \\ \hline
  $\boldsymbol{z}^{(i)}$       & State vector of a measurement point with index $i$. \\ \hline
  $(p_x, p_y, p_z)$  & Point object coordinate in Cartesian coordinate system. \\ \hline
  $(\gamma_k,\alpha_k, \beta_k)$  & Point object coordinate in sphere coordinate system. \\ \hline
  $\mathbb{M}, \mathbb{M}^f$       & The map space. Visible space in the map space. \\ \hline
  $\mathbb{V}_{i}, \mathbb{P}_{i}$       & Voxel subspace and pyramid subspace with index $i$.  \\ \hline
  $\mathbb{A}^{\boldsymbol{x}_k}$       & Activation space of a point object $\boldsymbol{x}_k$.  \\ \hline
  $\mathbb{A}^{\tilde{\boldsymbol x}_k^{(i)}}$  & Activation space of a particle $\tilde{\boldsymbol x}_k^{(i)}$.  \\ \hline
  $ D_{\text{X}}(\boldsymbol{x})$       & PHD of RFS $\text{X}$ at state $\boldsymbol{x}$. \\ \hline
  $\tilde{\boldsymbol x}_{k}^{(i)}, \tilde{\boldsymbol x}_{b,k}^{(i)}$      & The state vector of a particle and a newborn particle. \\ \hline
  $\tilde{\boldsymbol x}_{s,k|k-1}^{(i)}$ & The state vector of a particle survived from $k-1$ to $k$.\\ \hline
  $w^{(i)}_k$       & The weight of a particle with index $i$ at time $k$. \\ \hline
  $P_d, P_s$       & Detection and survival probability of an object. \\ \hline
  $N_k, M_k$ & Number of point objects and measurement points at $k$.  \\ \hline
  $N_v, N_p$ & Number of voxel subspaces and pyramid subspaces.  \\ \hline 
  $N_f$ & Number of pyramid subspaces in FOV.  \\ \hline
  $L_{k}$   & Number of particles at time $k$. \\ \hline
  $L_b$ & Number of newborn particles from a measurement point.  \\ \hline
  $L_{max}$ & Allowed max particle number in $\mathbb{M}$ after resampling.  \\ \hline
  $L_{max}^{\mathbb{V}}$ & Allowed max particle number in $\mathbb{V}_{i}$ after resampling.  \\ \hline
  $\lambda_1, \lambda_2$ & Coefficients in the mixture motion model. \\ \hline
  $\gamma_{k|k-1}, \kappa_k$       & Intensity of the newborn objects and clusters. \\ \hline
  $(l_x, l_y, l_z)$ & Size of the map space. \\ \hline
  $l$  & Side length or resolution of a voxel subspace. \\ \hline
  $n$  & The number of adjacent pyramids on each side in $\mathbb{A}^{\boldsymbol{x}_k}$. \\ \hline
  $Res$ & Resolution of the voxel filter for point cloud pre-process. \\ \hline
  $r_{min}$ & The radius of the robot sphere model. \\ \hline
  $(\theta_h, \theta_v)$ & Horizontal and vertical angle of the FOV. \\ \hline
  $\theta$  & Angle of a pyramid subspace. \\ \hline
  $\boldsymbol{Q}$  &  Gaussian noise covariance matrix in prediction step. \\ \hline
  $W^{(\mathbb V)}_d$ &  Weight sum of dynamic particles in a voxel subspace. \\ \hline
  $W^{(\mathbb V)}_s$ &  Weight sum of static particles in a voxel subspace. \\ \hline
  $W^{(\mathbb V)}_{d,s}$ &  Weight sum of all the particles in a voxel subspace. \\ \hline
  $V(\cdot)$ &  Function to calculate the absolute velocity value. \\ \hline
  $m(\cdot), pr(\cdot)$ &  Mass function and probability function in DST. \\ \hline
  $bel(\cdot), pl(\cdot)$ &  Belief function and plausibility function in DST. \\ \hline
  $\pi_{k|k-1}(\cdot)$   & State transition density function of a single object. \\ \hline
  $g_{k}(\cdot)$   & Measurement likelihood function of a single object. \\ \hline
  $f_Q(\cdot)$  & State transition function of a single point object. \\ \hline
  $f_R(\cdot)$  & Measurement function of a single point object. \\ \hline
  $R(\cdot)$  & Function that defines the measurement noise matrix. \\ \hline
  $\rho(\cdot)$  & Function that defines the standard deviation on each axis. \\ \hline
  \end{tabularx}
  \label{Table: Notation}
\end{table}

\section{Preliminaries} \label{Section: Preliminaries}
This section introduces the main concepts of RFS, PHD, PHD filter, and SMC-PHD filter. The relationship between the concepts is: PHD is a first moment of an RFS; PHD filter realizes multi-object tracking by propagating PHD; SMC-PHD filter is a particle-based implementation of the PHD filter and is used to fulfill prediction and update in our DSP map. The notations used in this section and the rest sections are shown in Table \ref{Table: Notation}.

\subsection{Random Finite Set}
An RFS is a finite set-valued random variable \cite{RFSMap}. The number and the states of the elements in an RFS are random but finite. 
Let $\text{X}$ denote an RFS and $\boldsymbol{x}^{(i)} \in \mathbb{M}$ denote the state vector of an element in $\text{X}$. $\mathbb{M}$ is $\boldsymbol{x}^{(i)}$'s state space, e.g., map space. Then $\text{X}$ is expressed as:
\begin{equation}\label{Eq: RFS intro}
  \text{X} = \left\{ \boldsymbol{x}^{(1)}, \boldsymbol{x}^{(2)}, ..., \boldsymbol{x}^{(N)} \right\}
\end{equation}
where $N \in \mathbb{N}$ is a random variable representing elements number in $\text{X}$ and is called the cardinality of $\text{X}$. Specially, when $N=0$, $\text{X}$ is $\emptyset$. A common usage of the RFS is in the multi-object tracking area, where $\boldsymbol{x}^{(i)}$ is usually the state of an object and $\text{X}$ is the set composed of the states of all objects. $N$ varies as objects appear and disappear in the tracking range.

\subsection{PHD}
PHD \cite{PHD2003,2013ParticleFilterBook} is a first moment of an RFS and is raised to describe the multi-object density. The PHD of $\text{X}$ at a state $\boldsymbol{x}$ is defined as: 
\begin{equation}\label{Eq: PHD}
  D_{\text{X}}(\boldsymbol{x}) = \mathbf{E} \left[ \sum_{\boldsymbol{x}^{(i)}\in \text{X}} \delta (\boldsymbol x-\boldsymbol{x}^{(i)}) \right] 
\end{equation}
where $\mathbf{E}[\cdot]$ is the expectation and $\delta(\cdot)$ is the Dirac function\footnote{Dirac function: $\delta(\boldsymbol{x})=0, \ \text{if} \ \boldsymbol x \neq \boldsymbol{0}$; $\int \delta(\boldsymbol{x}) \text d \boldsymbol{x}= 1$.}.

Two important properties of PHD are used in this work. The first property is that the integral of PHD is the expectation of the cardinality of $\text{X}$, which can be expressed as
\begin{equation}\label{Eq: PHD_integral}
  \int D_{\text{X}}(\boldsymbol{x}) \text{d}\boldsymbol x = \mathbf{E} [|\text{X}|]
\end{equation}
where $|\text{X}|$ represents the cardinality of $\text{X}$. 

Another property is that if $\text{X}^{(1)}, \text{X}^{(2)}, ..., \text{X}^{(N^\prime)}$ are independent RFSs, and $\text{X}^{(1)} \cup \text{X}^{(2)} \cup ... \cup \text{X}^{(N^\prime)} = \text{X}$, then
\begin{equation}\label{Eq: PHD_sum}
  D_\text{X}(\boldsymbol{x})= D_{\text{X}^{(1)}}(\boldsymbol{x}) + D_{\text{X}^{(2)}}(\boldsymbol{x}) + \cdots + D_{\text{X}^{(N^\prime)}}(\boldsymbol{x})
\end{equation}

\subsection{PHD Filter} 
The PHD filter \cite{PHD2003} is an efficient filter that propagates the PHD in the prediction and the update step, and can be used to handle multiple object tracking problems. Let $\text{X}_{k-1}$ and $\text{X}_{k}$ denote the RFS composed of object states at time step $k-1$ and $k$, respectively. Suppose $\text{Z}_k$ is the RFS composed of measurements, i.e., point cloud, to the objects at time $k$. 
In the prediction step of a typical PHD filter, the prior object states RFS $\text{X}_{k|k-1}$ can be treated as the union of two independent subsets, which is $\text{X}_{k|k-1} = \text{S}_{k|k-1} \cup \text{B}_{k|k-1}$, where $\text{S}_{k|k-1}$ represents the persistent objects from the $\text{X}_{k-1}$, and $\text{B}_{k|k-1}$ is the newly born objects. Note $\text{S}_{k|k-1}$ and $\text{B}_{k|k-1}$ are distinguished by birth time. They are both in map space $\mathbb{M}$ but don't share any element. $\text{S}_{k|k-1}$ is usually modeled with Multi-Bernoulli mixture (MBM). From $\text{X}_{k-1}$ to $\text{S}_{k|k-1}$, the objects have a probability of $P_s$ to survive. Meanwhile, $\text{B}_{k|k-1}$ is modeled as a Poisson point process (PPP) \cite{PHD2003} with intensity $\gamma_{k|k-1}(\boldsymbol{x}_k)$. 
Similarly, in the update step, $\text{Z}_k$ is expressed as $\text{Z}_k = \text{O}_k \cup \text{C}_k$, where $\text{O}_k$ is the detected objects set and $\text{C}_k$ is the set of clutter. From $\text{X}_k$ to $\text{Z}_k$, the objects have a probability of $P_d$ to be detected. The clutter $\text{C}_k$ are modeled as a PPP with intensity $\kappa_{k}(\boldsymbol{z}_k)$.

Let $D_{\text{S}_{k|k-1}}(\boldsymbol{x}_k)$ and $D_{\text{B}_{k|k-1}}(\boldsymbol{x}_k)$ denote the PHD at $\boldsymbol{x}_k$ of RFS $\text{S}_{k|k-1}$ and $\text{B}_{k|k-1}$, respectively. Considering the property (\ref{Eq: PHD_sum}) and the MBM and PPP models, the general PHD filter \cite{2013ParticleFilterBook} is described as:
\begin{equation}\label{Eq: PHD_filter_prediction1}
\begin{aligned}
  & D_{\text{X}_{k|k-1}}(\boldsymbol{x}_k) = D_{\text{S}_{k|k-1}}(\boldsymbol{x}_k) + D_{\text{B}_{k|k-1}}(\boldsymbol{x}_k) \qquad \quad \ \\
  &  \qquad \qquad \quad \   = P_s H_k(\boldsymbol{x}_k, \boldsymbol{x}_{k-1}) + \gamma_{k|k-1}(\boldsymbol{x}_k)  \\
\end{aligned}
\end{equation}
\begin{equation}
\label{Eq: PHD_filter_prediction2}
   \small H_k(\boldsymbol{x}_k, \boldsymbol{x}_{k-1}) =  \int \pi_{k|k-1} (\boldsymbol{x}_k|\boldsymbol{x}_{k-1}) D_{\text{X}_{k-1}}(\boldsymbol{x}_{k-1}) \text{d}\boldsymbol{x}_{k-1}
\end{equation}
{\small
\begin{align}
  \label{Eq: PHD_filter_update1}
  & D_{\text{X}_{k}}(\boldsymbol{x}_k) = \left[ 1-P_d + P_d \sum_{\boldsymbol{z}_k\in \text{Z}_k} G_k(\boldsymbol{z}_k, \boldsymbol{x}_k)\right]D_{\text{X}_{k|k-1}}(\boldsymbol{x}_k) \\
  \label{Eq: PHD_filter_update2}
  & G_k(\boldsymbol{z}_k, \boldsymbol{x}_k) = \frac{ g_k(\boldsymbol{z}_k|\boldsymbol{x}_k)}{\kappa_k (\boldsymbol{z}_k) + P_d \int g_k(\boldsymbol{z}_k|\boldsymbol{x}_k)D_{\text{X}_{k|k-1}}(\boldsymbol{x}_k)\text{d}\boldsymbol{x}_k }
\end{align}}where Equation (\ref{Eq: PHD_filter_prediction1}) and (\ref{Eq: PHD_filter_prediction2}) show the prediction step, and Equation (\ref{Eq: PHD_filter_update1}) and (\ref{Eq: PHD_filter_update2}) present the update step. $\pi_{k|k-1}(\cdot)$ is the state transition density of a single object and $g_k(\cdot)$ is the single object measurement likelihood.

\subsection{SMC-PHD Filter} \label{Section: SMC-PHD filter}
Sequential Monte Carlo PHD (SMC-PHD) filter \cite{SMC-PHD2005} \cite{ImprovedSMC2010} uses particles to represent PHD and is an efficient implementation of the PHD filter. Each particle has a weight and a state vector with the same dimension as an object's state. With the particles, the posterior PHD of $\text{X}$ at time $k-1$ is approximated by
\begin{equation}\label{Eq: PHD particles}
  D_{\text{X}_{k-1}}(\boldsymbol{x}_{k-1}) \approx \sum_{i=1}^{L_{k-1}} w_{k-1}^{(i)} \delta(\boldsymbol{x}_{k-1} - \tilde{\boldsymbol x}_{k-1}^{(i)})
\end{equation}
where $L_{k-1}$ is the number of particles at time step $k-1$, $w_{k-1}^{(i)}$ is the weight of particle with index $(i)$, and $\tilde{\boldsymbol x}_{k-1}^{(i)}$ denotes the state vector of particle $(i)$. We distinguish the state of an object and the state of a particle with the tilde notation.

In the prediction step, with Equations (\ref{Eq: PHD particles}), (\ref{Eq: PHD_filter_prediction1}) and (\ref{Eq: PHD_filter_prediction2}), the prior PHD of the RFS $\text{X}_{k|k-1}$ at $k$ is derived as:
\begin{equation}\label{Eq: prediction SMC-PHD}
\begin{aligned}
  & D_{\text{X}_{k|k-1}}(\boldsymbol{x}_k) = D_{\text{S}_{k|k-1}}(\boldsymbol{x}_k) + D_{\text{B}_{k|k-1}}(\boldsymbol{x}_k) \\
  & \qquad  =  \sum_{i=1}^{L_{k-1}}P_s w_{k-1}^{(i)}  \pi_{k|k-1}(\boldsymbol{x}_k|\tilde{\boldsymbol x}_{k-1}^{(i)}) + \gamma_{k|k-1}(\boldsymbol{x}_k) \\
\end{aligned}
\end{equation}

Let $w_{s,k|k-1}^{(i)} = P_s w_{k-1}^{(i)}$. By sampling $\pi_{k|k-1}(\boldsymbol{x}_k|\tilde{\boldsymbol x}_{k-1}^{(i)})$ and $\gamma_{k|k-1}(\boldsymbol{x}_k)$ with particles, the above equation can be further derived as:
\begin{equation} \label{Eq: prediction SMC-PHD Survived 211}
  \small
  \begin{aligned}
    & D_{\text{X}_{k|k-1}}(\boldsymbol{x}_k) \\
    & = \sum_{i=1}^{L_{k-1}} w_{s,k|k-1}^{(i)} \delta(\boldsymbol{x}_k - \tilde{\boldsymbol x}_{s,k|k-1}^{(i)}) +
        \sum_{j=1}^{L_{b,k}} w_{b,k}^{(i)} \delta(\boldsymbol{x}_k - \tilde{\boldsymbol x}_{b,k}^{(j)}) \\
    &  \equiv \sum_{i=1}^{L_{k}} w_{k|k-1}^{(i)} \delta(\boldsymbol{x}_k - \tilde{\boldsymbol x}_{k}^{(i)})
  \end{aligned}
\end{equation}
where $\tilde{\boldsymbol x}_{s,k|k-1}^{(i)}$ represents the particle state sampled from $\pi_{k|k-1}(\boldsymbol{x}_k|\tilde{\boldsymbol x}_{k-1}^{(i)})$ and $\tilde{\boldsymbol x}_{b,k}^{(j)}$ represents the particle state sampled from $\gamma_{k|k-1}(\boldsymbol{x}_k)$. $L_{b,k}$ and $w_{b,k}^{(i)}$ are the number and the weight of newborn particles at time $k$, respectively. The total number of particles after prediction is $L_{k}=L_{k-1}+L_{b,k}$.

In the update step, substitute $D_{\text{X}_{k-1}}(\boldsymbol{x}_{k-1})$ in Equations (\ref{Eq: PHD_filter_update1}) and (\ref{Eq: PHD_filter_update2}) with the particle representation in the last row of (\ref{Eq: prediction SMC-PHD Survived 211}). The posterior PHD at $k$ is reformed into the summation of particles, which is
\begin{equation}\label{Eq: PHD particles_posterior}
  D_{\text{X}_{k}}(\boldsymbol{x}_k) \approx \sum_{i=1}^{L_{k}} w_{k}^{(i)} \delta(\boldsymbol{x}_k - \tilde{\boldsymbol x}_{k}^{(i)})
\end{equation}
where the particle state $\tilde{\boldsymbol x}_{k}^{(i)}$ remains the same as in the prediction step and the weight $w_k^{(i)}$ is given by:
\begin{align}
  \label{Eq: particles_posterior_weights1}
  w_k^{(i)} = \left[ 1-P_d + \sum_{\boldsymbol{z}_k \in \text{Z}_k} \frac{P_d g_{k}(\boldsymbol{z}_k|\tilde{\boldsymbol x}_{k}^{(i)})}{ \kappa_k(\boldsymbol{z}_k) + \text{C}_k(\boldsymbol{z}_k) } \right] & w_{k|k-1}^{(i)} \\
  \label{Eq: particles_posterior_weights2}
  \text{C}_k(\boldsymbol{z}_k) = \sum_{j=1}^{L_k} P_d w_{k|k-1}^{(j)} g_{k}(\boldsymbol{z}_k|\tilde{\boldsymbol x}_{k}^{(j)}) &
\end{align}


The SMC-PHD filter estimates the PHD of $\text{X}$ by iterative prediction with Equation (\ref{Eq: PHD particles}) to (\ref{Eq: prediction SMC-PHD Survived 211}) and update with Equation (\ref{Eq: PHD particles_posterior}) to (\ref{Eq: particles_posterior_weights2}). Details can be found in \cite{SMC-PHD2005} \cite{ImprovedSMC2010}.

\section{World Model and System Overview} \label{Section: world model}
\subsection{World Model}
\begin{figure*}
\centerline{\psfig{figure=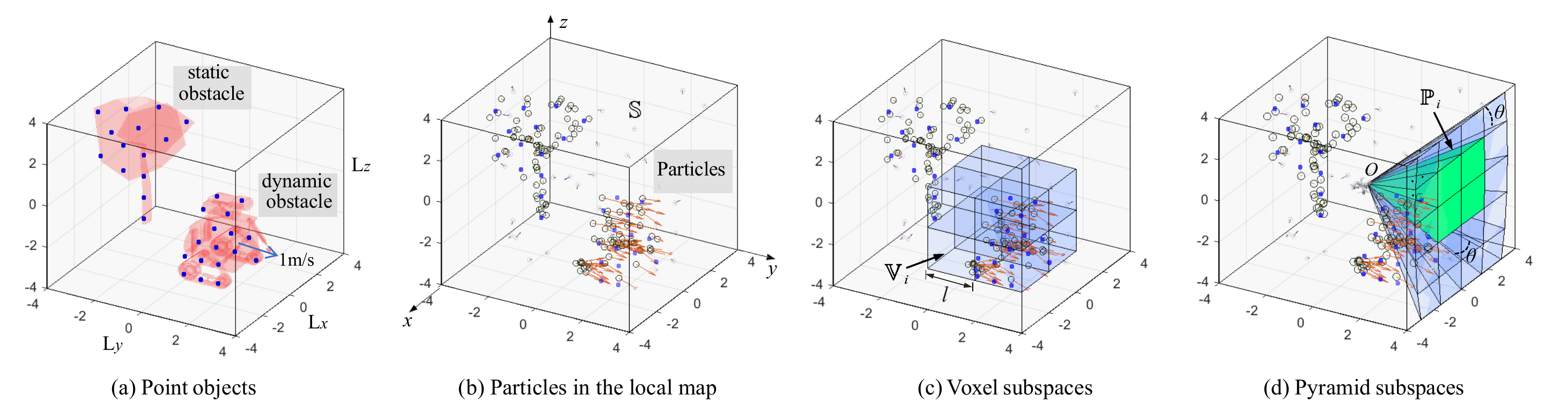,width=7.2in}}
\caption{Illustration of the world model. Subfigure (a) shows a cubic local environment with a static obstacle and a dynamic obstacle. The small blue points are the point objects that represent the two obstacles. The layout of the point objects depends on the measurement points. The blue points show one kind of layout. Subfigure (b) presents the particles (small hollow circles with arrows indicating the velocities) used to model the point objects. Subfigure (c) and Subfigure (d) are two different space division structures. The whole local environment is divided into subspaces, but only a part of the subspaces are plotted to have a clear view of their shapes. In Subfigure (d), the green pyramids indicate the current FOV.}
\label{Fig: dual structure}
\end{figure*}

Our DSP map is an egocentric map built on multi-object tracking at the point object level in a continuous neighborhood space. 
Let $\mathbb{M}$ denote the neighborhood map space of the robot. $\mathbb{M}$ is a real space that has a cuboid boundary with size $(l_x, l_y, l_z)$. The size can be set according to the range of the utilized sensors or the requirements from the motion planner. At the center of the cuboid is the robot.
We consider the obstacles in $\mathbb{M}$ as point objects, similar to \cite{RFSMap}. Fig. \ref{Fig: dual structure}(a) reveals the relation between obstacles and point objects. One obstacle can correspond to multiple point objects. 
The point objects are used to estimate the occupancy status at an arbitrary position in the map. Since the occupancy status rather than the state of each obstacle is more important in an occupancy map, the mapping from point objects to obstacles is omitted and the assumption that all the point objects move independently is made.
The same assumption is used in the existing works on particle-based maps \cite{SMCPHDMap} \cite{RFSMap} \cite{DynamicMapICRA2021}.

%
%
%
%
For the reason that the obstacles are unknown, the number of the point objects in $\mathbb{M}$ and their states are random but finite. Therefore, these point objects can be modeled as an RFS. At a discrete time $k$, the RFS composed of the point object states is represented as
\begin{equation}\label{Eq: RFS}
  \text{X}_k = \left\{ \boldsymbol{x}^{(1)}, \boldsymbol{x}^{(2)}, ..., \boldsymbol{x}^{(N_k)} \right\}
\end{equation}
where $N_k$ is the number of point objects at time $k$, and $\boldsymbol x$ with index from $1$ to $N_k$ is the state vector of a point object. The state vector is given by the 3D position and velocity, namely
\begin{equation}\label{Eq: State}
  \boldsymbol{x}=\left[ p_{x}, p_{y}, p_{z}, v_{x}, v_{y}, v_{z} \right] ^T
\end{equation}
where the subscripts $\{{x},{y},{z}\}$ are used to represent the axes in Cartesian coordinate.  The core of building the DSP map is to use the SMC-PHD filter to track the point objects in 3D continuous space and estimate $\text{X}_k$'s PHD, which is then used to estimate the occupancy status of the map.

\begin{figure}
\centerline{\psfig{figure=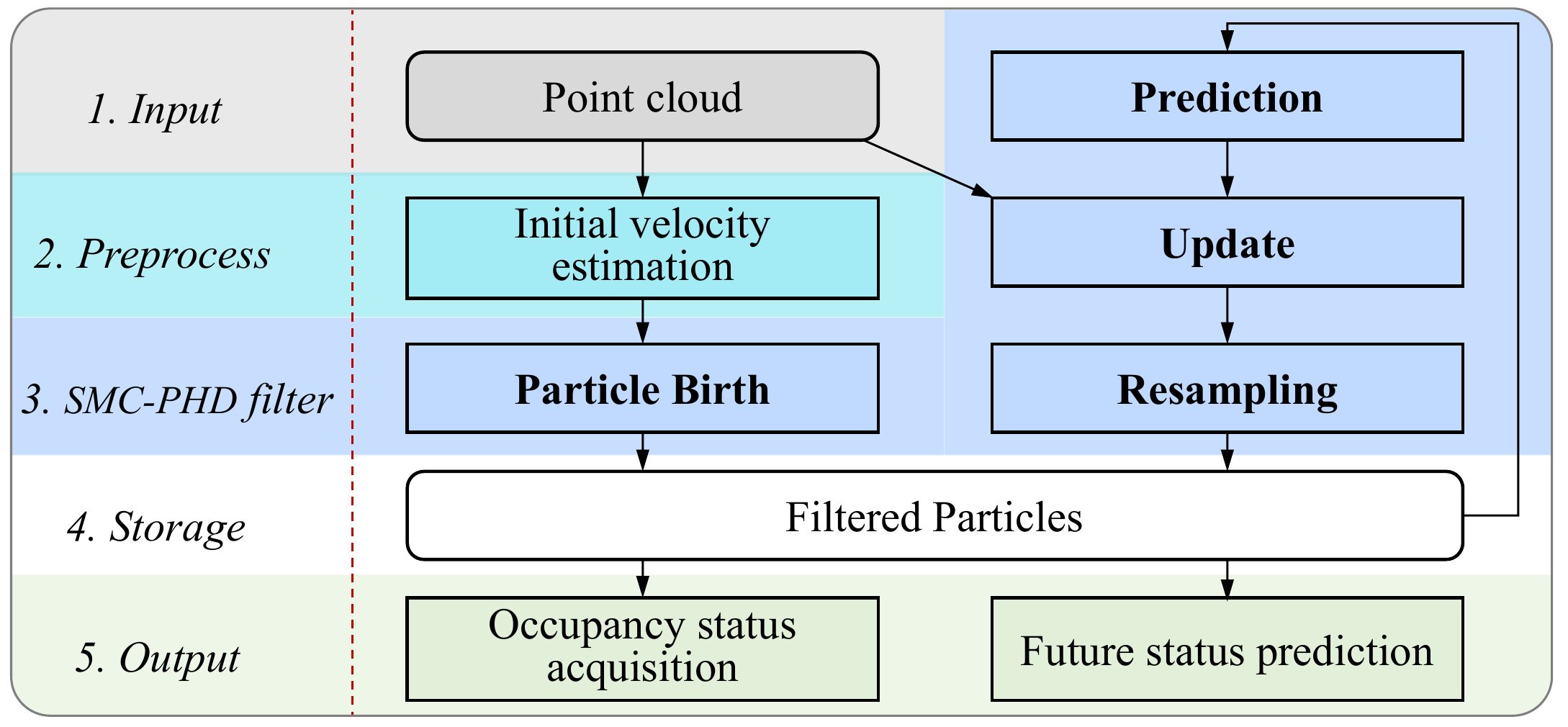, width=3.4in}}
  \caption{System overview of our DSP map.}
  \label{Fig: system}
\end{figure}

To realize effective and efficient SMC-PHD filtering in the continuous space, we divide $\mathbb M$ into two types of subspaces, i.e., the cubic voxel subspaces and the pyramid-like subspaces, by the position dimensions. The voxel subspaces are used for data storage and particle resampling. The pyramid-like subspaces are applied to handle limited sensor FOV and inevitable occlusions in the continuous space, and realize efficient particle update. Details are presented in Section \ref{Section: Map Building}. The following describes how to acquire the subspaces and defines the sub-RFSs divided accordingly with the subspaces.



The voxel subspaces are divided in the cartesian coordinate (Fig. \ref{Fig: dual structure}(c)).
The voxels can fill up $\mathbb{M}$ but have no overlaps with each other. Assume the resolution of the voxel is $l$. 
Then the number of the voxels is $N_v = \frac{l_x \cdot l_y \cdot l_z}{l^3}$. Let $\mathbb{V}_{i}$ denote the $i^{th}$ voxel subspace.  
Then $\text{X}_k$ can be described as the union of these sub-RFSs, which is
\begin{equation}\label{Eq: RFS_Union_V}
  \text{X}_k = \text{X}_k^{(\mathbb{V}_1)} \cup \text{X}_k^{(\mathbb{V}_2)} \cup \cdots \cup \text{X}_k^{(\mathbb{V}_{N_v})}
\end{equation}
Since the voxels have no overlaps, any two sub-RFSs don't share a point object, and thus, the sub-RFSs are independent.
In the SMC-PHD filter, the voxel subspaces are used to resample the particles in $\mathbb{M}$ in a uniform manner, which is described in Section \ref{Section: Resampling}. In addition, these voxel subspaces are used to index and store the particles for efficiency purposes, as described in Section \ref{Section: Implementation}.  

For the reason that the field of view (FOV) of a sensor is usually limited, and the occlusion prevents observations of the area behind obstacles, only a part of $\mathbb{M}$ is visible. Let $\mathbb{M}^f \subset \mathbb{M}$ denote the visible space. $\mathbb{M}^f$ must be distinguished from the occluded space to realize map updating.
However, the voxel subspaces have a limited resolution and cannot continuously express $\mathbb{M}^f$.
Thus, another division structure is still required.


Inspired by the perspective projection model for sensors, we also divide $\mathbb{M}$ into pyramid-like subspaces in the spherical coordinate (Fig. \ref{Fig: dual structure}(d)). These subspaces are divided dynamically and uniformly in the sensor frame when the robot pose is given. (Details can be found in Section \ref{Section: Implementation} and Algorithm \ref{Algorithm: Pyramid Particle} in the Appendix.) The real shape of a pyramid-like subspace is composed of four near-triangular faces and one face on the map boundary face. For simplification, we loosely name the subspace as pyramid subspace in the following content.

In the spherical coordinate, the azimuth angle range is $\left[ 0, 2 \pi \right]$ and the zenith angle range is $\left[ 0, \pi\right]$. Suppose the angle interval of the pyramid division, namely the pyramid angle, is $\theta > 0$. 
The number of these subspaces is $N_p = \frac{2\pi \cdot \pi}{\theta^2}$. To make $N_p$ an integer, $\theta$ satisfies $I \theta = \pi$, where $I \in \mathbb{N}+$.
Denote by $\mathbb{P}_{i}$ the $i^{th}$ pyramid subspace, and by $\text{X}_k^{(\mathbb{P}_i)}$ the RFS composed of point objects in $\mathbb{P}_{i}$.
$\text{X}_k$ satisfies
\begin{equation}\label{Eq: RFS_Union_A}
  \text{X}_k = \text{X}_k^{(\mathbb{P}_1)} \cup \text{X}_k^{(\mathbb{P}_2)} \cup \cdots \cup \text{X}_k^{(\mathbb{P}_{N_p})}
\end{equation}

The measurement of the point objects is the point cloud from sensors, such as stereo cameras or Lidars. 
The points in the point cloud at time $k$ form a measurement RFS $\text{Z}_k$. In analogy to the point objects, $\text{Z}_k$ is written as
\begin{equation}\label{Eq: RFS Measurement}
  \text{Z}_k = \left\{ \boldsymbol{z}^{(1)}, \boldsymbol{z}^{(2)}, ..., \boldsymbol{z}^{(M_k)} \right\}
\end{equation}
where $M_k$ represents the number of the measurement points, and each measurement point $\boldsymbol{z}$ consists of the 3D position, which is
\begin{equation}\label{Eq: point_z}
  \boldsymbol{z} = \left\{ z_{x}, z_{y}, z_{z} \right\}
\end{equation}

\begin{figure*}
  \centerline{\psfig{figure=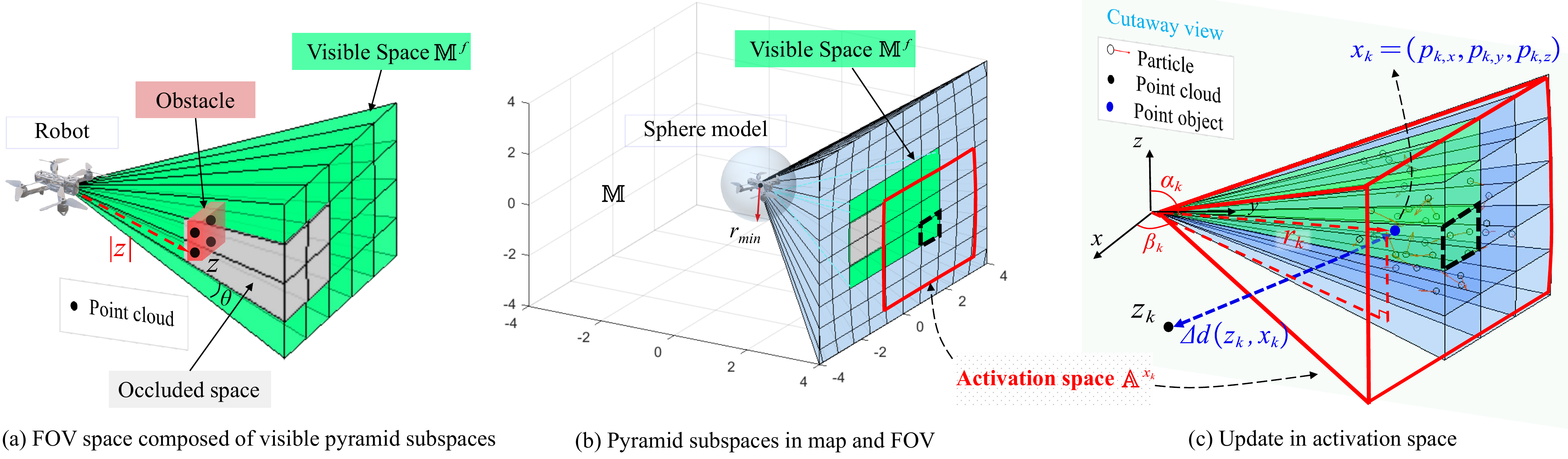, width=7.0in}}
  \caption{Illustration of the pyramid subspaces in the FOV and the update step. Subplot (a) shows the visible space and the occluded space. Subplot (b) reveals the visible space and the map space in the same plot. The pyramid subspaces out of the current FOV are partially presented to have a clear view. For a point object $x$ that lies in a pyramid subspace belonging to the visible space, such as the subspace outlined with black dashes, we define the activation space of $x$ as the union of this pyramid subspace and its adjacent $n$ pyramid subspaces. The adjacent $n$ pyramid subspaces of a pyramid are within the range of $n$ rows and $n$ columns from the pyramid. 
  $n=2$ in this case. Subplot (c) shows a cutaway view of the activation space. Point cloud measurement $\boldsymbol{z}_k$ locates out of the activation space. The distance from the point object $x$ to the map center is $r_k$.}
  \label{Fig: update}
\end{figure*}

With the measurement points and the pyramid-like subspaces, we can determine the visible space $\mathbb{M}^f$ and occluded space.  
As is shown in green in Fig. \ref{Fig: update}. (a) and (b), $\mathbb{M}^f$ is the union of the free space and obstacle surface in each pyramid subspace in the FOV.
Denote the visible space of pyramid subspace $\mathbb{P}_{i}$ by $\mathbb{P}_{i}^f$.
When the pyramid angle $\theta$ of $\mathbb{P}_{i}$ equals the angular resolution of the sensor, there is either one or no measurement point in $\mathbb{P}_{i}$. If there is one point $\boldsymbol{z}$, the subspace behind the measurement point is occluded (painted in gray in Fig. \ref{Fig: update}), while the rest space is the visible pyramid subspace $\mathbb{P}_{i}^f$. $\mathbb{P}_{i}^f \subset \mathbb{P}_{i}$ and the length of $\mathbb{P}_{i}^f$ is $|\boldsymbol{z}|$.
If there is no measurement point, $\mathbb{P}_{i}^f = \mathbb{P}_{i}$.
Suppose the FOV is $\theta_h \times \theta_v$. The number of $\mathbb{P}_{i}^f$ is $N_f = \frac{\theta_h \theta_v}{\theta^2}$. Since the FOV usually cannot cover the whole neighborhood space, $N_f < N_p$. Then $\mathbb{M}^f = \mathbb{P}_{1}^f \cup \cdots  \cup \mathbb{P}_{N_f}^f $, and $\text{Z}_k$ can be divided into subsets with these visible pyramid subspaces, which is
\begin{equation}\label{Eq: RFS_Union_Z}
  \text{Z}_k = \text{Z}_k^{(\mathbb{P}_1^f)} \cup \text{Z}_k^{(\mathbb{P}_2^f)} \cup \cdots \cup \text{Z}_k^{(\mathbb{P}^f_{N_f})}
\end{equation}

With the measurement $\text{Z}_k$, the PHD of $\text{X}_k$ is updated by using the SMC-PHD filter. The hollow circles with velocity arrows in Subfigures (b), (c), and (d) in Fig. \ref{Fig: dual structure} show the particles used in the SMC-PHD filter. The basic element in our map is the particle. 

\subsection{System Overview}
An overview of the procedures to build our DSP map can be found in Fig. \ref{Fig: system}. The core procedure is filtering the PHD of $\text{X}_k$ with the SMC-PHD filter. In the SMC-PHD filter, the prediction step and update step iteratively update the PHD. The particle birth step generates new particles and is then used in the prediction step. A resampling step is added after the update step to prevent degeneration and control the maximum number of particles. The pyramid subspace division is used in the update step to distinguish visible space and improve computational efficiency. The voxel subspace division is used in the resampling step to realize efficient and uniform particle resampling.
Details about the SMC-PHD filtering procedure in our map can be found in Section \ref{Section: Map Building}.
The filtered result is particles with position and velocity states. Then the map output can be calculated with the particles into two forms designed for motion planning. The first form is the current occupancy status, and the second is the prediction of future occupancy status.
An initial velocity estimation procedure is introduced to reduce the noise in mapping.
Details about the output and initial velocity estimation are presented in Section \ref{Section: extensions in mapping}.
The particles used in all the procedures are stored in the voxel subspaces. 




\section{Mapping with Dual Structure} \label{Section: Map Building}
This section presents the core procedures to build our map with the dual-structure space divisions, including prediction, update, particle birth, and resampling steps. The prediction and the particle birth are conducted in space $\mathbb{M}$. In the update step, the pyramid subspaces $\mathbb{P}_{i}$ are utilized to update the point objects' PHD efficiently. The voxel subspaces $\mathbb{V}_{i}$ are adopted in the resampling step.

\subsection{Prediction} \label{Section: Prediction}
The prediction step predict the prior PHD of $\text{X}_{k|k-1}$ and the general form has been described in Section \ref{Section: SMC-PHD filter}. In our map, the motion model of a single point object is defined by the constant velocity (CV) model, then a point object $\boldsymbol{x}_k$ that survived from $k-1$ is predicted by:
\begin{equation}\label{Eq: single_object_motion}
 \boldsymbol{x}_k = f_Q \left( \boldsymbol{x}_{k-1} \right) + \boldsymbol{\xi}
 =  \left[\begin{array}{cc}
\boldsymbol I_{3 \times 3} & \Delta t \boldsymbol I_{3 \times 3} \\
0_{3 \times 3} & \boldsymbol I_{3 \times 3}
\end{array}\right] \boldsymbol{x}_{k-1} + \boldsymbol{\xi}
\end{equation}
where $\boldsymbol I$ is the identity matrix and $\boldsymbol{\xi}$ is the noise. The noise is supposed to obey a Gaussian distribution with a covariance $\boldsymbol{Q}$, which is $\boldsymbol{\xi} \sim \mathcal{N}(0,\boldsymbol{Q})$.

Then the state transition density in Equation (\ref{Eq: PHD_filter_prediction2}) turns to a Gaussian probability density:
\begin{equation}\label{Eq: state transition density}
  \pi_{k|k-1} (\boldsymbol{x}_k|\boldsymbol{x}_{k-1}) = \mathcal{N}\left(\boldsymbol{x}_k ; f_{Q}\left(\boldsymbol{x}_{k-1} \right), \boldsymbol{Q}\right)
\end{equation}
Thus, from Equation (\ref{Eq: prediction SMC-PHD}) to (\ref{Eq: prediction SMC-PHD Survived 211}), $\pi_{k|k-1}(\boldsymbol{x}_k|\tilde{\boldsymbol x}_{k-1}^{(i)})$ can be sampled by particles using the Gaussian probability density, and $\tilde{\boldsymbol x}_{s,k|k-1}^{(i)}$ in Equation (\ref{Eq: prediction SMC-PHD Survived 211}) is given by
\begin{equation}\label{Eq: particle prediction cv}
  \tilde{\boldsymbol x}_{s,k|k-1}^{(i)} = f_Q \left( \tilde{\boldsymbol x}_{k-1}^{(i)} \right) + \boldsymbol{u}
\end{equation}
where $\boldsymbol{u}$ is a noise sampled from $\mathcal{N}(0, \boldsymbol{Q})$. 

The weight and state of newborn particles in Equation (\ref{Eq: prediction SMC-PHD Survived 211}) are described later in Section \ref{Section: Newborn Particles}. 


\subsection{Update} \label{Section: Update}
The update step utilizes the measurement $\text{Z}_k$ to get the posterior PHD of $\text{X}_k$. Two major points are addressed in the update step. The first point is to tackle the limited FOV and the occlusion.
Since the FOV of a sensor is usually limited, and the occlusion prevents observations to the area behind obstacles, $\text{Z}_k$ can only be in the visible space $\mathbb{M}^{f}$ defined in Section \ref{Section: world model}.
The objects that do not belong to $\mathbb{M}^{f}$ are in an unknown area and should not be updated; otherwise, their existence probability will be falsely reduced. Thus, the notations in Equation (\ref{Eq: PHD particles_posterior}) should contain a superscript ${f}$, such as $D^{f}_{\text{X}_{k}}(\boldsymbol{x}_k)$, $w_{k}^{f, (i)}$ and $\tilde{\boldsymbol x}_{k}^{f, (i)}$. Let superscript $\bar{f}$ represent the definitions in $\mathbb{M} \setminus \mathbb{M}^{f}$. Then by using the property in (\ref{Eq: PHD_sum}), the PHD of the all objects in $\mathbb{M}$ should be estimated as:
\begin{equation}\label{Eq: PHD particles_posterior_whole}
  D_{\text{X}_{k}}(\boldsymbol{x}_k) = D^{f}_{\text{X}_{k}}(\boldsymbol{x}_k) + D^{\bar{f}}_{\text{X}_{k}}(\boldsymbol{x}_k)
\end{equation}
where $D^{\bar{f}}_{\text{X}_{k}}(\boldsymbol{x}_k) = D^{\bar{f}}_{\text{X}_{k|k-1}}(\boldsymbol{x}_k)$ because the objects in $\mathbb{M} \setminus \mathbb{M}^{f}$ are not updated. 
$D^{f}_{\text{X}_{k}}(\boldsymbol{x}_k)$ can be updated with Equation (\ref{Eq: PHD particles_posterior}), (\ref{Eq: particles_posterior_weights1}) and (\ref{Eq: particles_posterior_weights2}) by considering the point objects and particles in $\mathbb{M}^{f}$ only. 
For notation simplification, the superscript ${f}$ is omitted in what following. We still use Equation (\ref{Eq: PHD particles_posterior}), (\ref{Eq: particles_posterior_weights1}) and (\ref{Eq: particles_posterior_weights2}) to represent the general update form but now $\tilde{\boldsymbol x}_{k}^{(i)} \in \mathbb{M}^{f}$ and $L_k$ only counts the particles in $\mathbb{M}^{f}$.



The second point is to reduce the computational complexity.
It should be noted that $\text{C}_k(\boldsymbol{z}_k)$ in Equation (\ref{Eq: particles_posterior_weights1}) and (\ref{Eq: particles_posterior_weights2}) is controlled by $\boldsymbol{z}_k$, and thus, for every $w_k^{(i)}$, $\text{C}_k(\boldsymbol{z}_k)$ can be shared for the same $\boldsymbol{z}_k$. Therefore, to calculate all the required $\text{C}_k(\boldsymbol{z}_k)$, the multiplication operation and PDF calculation calculation in (\ref{Eq: particles_posterior_weights2}) should be performed $L_k \cdot M_k$ times, where $M_k$ is the cardinality of $\text{Z}_k$. In addition, considering the summation operations in Equation (\ref{Eq: PHD particles_posterior}) and (\ref{Eq: particles_posterior_weights1}), another $L_k \cdot M_k$ times of multiplication, division and PDF calculation operations should be performed. The algorithmic complexity is $O(L_k M_k)$. In an unknown environment, there could be many obstacles and over a million particles can be required to approximate the states of the point objects. Hence, $L_k \cdot M_k$ can be very large, and the efficiency of the map is not adequate. The following considers using the pyramid subspaces to reduce the complexity.

Considering the measurement noise of the commonly used point cloud sensors, such as depth camera and Lidar, the single object measurement likelihood $g_{k}(\cdot)$ can be assumed as a Gaussian distribution, which is
\begin{equation}\label{Eq: likelihood}
  g_{k}(\boldsymbol{z}_k|\boldsymbol{x}_k) = \mathcal{N}(\boldsymbol{z}_k; f_R(\boldsymbol{x}_k), R(\boldsymbol{x}_k))
\end{equation}
where $f_R(\boldsymbol{x}_k) = \left[ \boldsymbol I_{3 \times 3}, 0_{3 \times 3} \right] \cdot \boldsymbol{x}_k$ since the measurement is only position.
Unlike the prediction covariance $\boldsymbol{Q}$, the measurement covariance $R(\boldsymbol{x}_k)$ is usually not constant but related to the distance $d_k$ of the obstacle in regular sensor models.  

Firstly, we assume that the measurement error of a point object $\boldsymbol{x}_k \in \mathbb{M}^f$ is independent on each axis, and the standard deviation on each axis is equally $\rho(d_k)$, where $\rho(\cdot)$ is a function. The coordinate of $\boldsymbol{x}_k$ in the sphere coordinate system is $(r_k, \alpha_k, \beta_k)$ (Fig. \ref{Fig: update} (c)), where $r_k=\sqrt{p_{k,x}^2+p_{k,y}^2+p_{k,z}^2}$, $\alpha_k=\arccos\frac{p_{k,z}}{r_k}$, and $\beta_k=\arccos \frac{p_{k,x}}{r\sin(\alpha_k)}$, if $p_{k,y} \geq 0$, and $\beta_k=\arccos\frac{p_{k,x}}{r\sin(\alpha_k)} + \pi$, if $p_{k,y}<0$. 

$r_k$ and $\alpha_k$ are bounded. 
Suppose the robot is in a sphere model with radius $r_{min} > 0$. The space inside the sphere is considered free and not updated. 
Let $r_{max} =\frac{1}{2} \sqrt{l_x^2+l_y^2+l_z^2}$. Then $r_k \in \left[ r_{min},\ r_{max} \right]$. Considering the sensor has a limited FOV with vertical view angle $\theta_v < \pi$, we have $\alpha_k \in \left[ \frac{\pi-\theta_v}{2}, \frac{\pi+\theta_v}{2} \right]$. 

With the coordinates in the sphere coordinate system, the covariance turns to $R(\boldsymbol{x}_k)= \rho^2 (r_k) \boldsymbol I_{3 \times 3}$ and $g_{k}(\boldsymbol{z}_k|\boldsymbol{x}_k)$ can then be rewritten as
\begin{equation}\label{Eq: likelihood_independent}
g_{k}(\boldsymbol{z}_k|\boldsymbol{x}_k) = \prod_{i \in \{{x},{y},{z} \}^{}} \mathcal{N}\left(z_{k,i}; p_{k,i}, \rho^2 (r_k) \right)
\end{equation}
where $z_{k, i}$ and $p_{k, i}$ are the single-axis position of a measurement and an object, respectively, at time $k$, which are described in (\ref{Eq: point_z}) and  (\ref{Eq: State}).

In Equation (\ref{Eq: particles_posterior_weights2}), $g_{k}(\boldsymbol{z}_k|\boldsymbol{x}_k)$ involves the transition density from $\boldsymbol{x}_k$ to every observed $\boldsymbol{z}_k$ and thus makes the update step time-consuming. 
Let $\epsilon$ denote a small constant. If $\boldsymbol{z}_k$ satisfies Condition $g_{k}(\boldsymbol{z}_k|\boldsymbol{x}_k) < \epsilon$, the approximation $g_{k}(\boldsymbol{z}_k|\boldsymbol{x}_k) = 0$, i.e. the strategy that the $\boldsymbol{z}_k$ is omitted in (\ref{Eq: particles_posterior_weights2}), is applied to improve the update efficiency. To find the $\boldsymbol{z}_k$ that satisfies the condition, we define an activation space $\mathbb{A}^{\boldsymbol{x}_k}$ for point object $\boldsymbol{x}_k$. The activation space indicates that $\boldsymbol{z}_k$ out of $\mathbb{A}^{\boldsymbol{x}_k}$ satisfies $g_{k}(\boldsymbol{z}_k|\boldsymbol{x}_k) \approx 0$.
$\mathbb{A}^{\boldsymbol{x}_k}$ is the adjacent space of $\boldsymbol{x}_k$, and is composed of the union of the pyramid subspace where $\boldsymbol{x}_k$ is and the adjacent $n$ pyramid subspaces. The adjacent $n$ pyramid subspaces are within the range of $n$ rows and $n$ columns from the pyramid subspace. For example, Fig. \ref{Fig: update}. (b) and (c) show the activation space of $\boldsymbol{x}_k$ with $n=2$. The number of pyramid subspaces in $\mathbb{A}^{\boldsymbol{x}_k}$ is $(2n+1)^2$. 

Let $\Delta d(\boldsymbol{z}_k,\boldsymbol{x}_k)$ denote the Euclidean position distance between an measurement point $\boldsymbol{z}_k$ and the point object $\boldsymbol{x}_k$. Consider the 3d Gaussian probability density, $g_{k}(\boldsymbol{z}_k|\boldsymbol{x}_k) $ can be further written as:
\begin{equation}\label{Eq: likelihood_independent_specified}
  g_{k}(\boldsymbol{z}_k|\boldsymbol{x}_k) = \frac{1}{(2\pi)^\frac{3}{2} \rho^3(r_k)} \ e^{-\frac{\Delta d(\boldsymbol{z}_k,\boldsymbol{x}_k)^2}{2 \rho^2(r_k)}}
\end{equation}

If $\boldsymbol{z}_k \notin \mathbb{A}^{\boldsymbol{x}_k}$, the absolute azimuth angle and the zenith angle difference between $\boldsymbol{x}_k$ and $\boldsymbol{z}_k$ is no less than $n\theta$. Let $\theta^\prime = n\theta$. 
In Appendix \ref{App: Lower Bound Calculation}, we derive that the lower bound of $\Delta d(\boldsymbol{z}_k,\boldsymbol{x}_k)$ is $r_{k} \sin \theta^\prime \sin \alpha_k$.
Thus for $\forall \boldsymbol{z}_k \notin \mathbb{A}^{\boldsymbol{x}_k}$, the maximum density is
\begin{equation}\label{Eq: likelihood_max}
  g_{k,max}(\boldsymbol{z}_k|\boldsymbol{x}_k) = \frac{1}{(2\pi)^\frac{3}{2} \rho^3(r_k)} \ e^{-\frac{(r_{k} \sin \theta^\prime \sin \alpha_k)^2}{2 \rho^2(r_k)}}
\end{equation}
When $(r_{k} \sin \theta^\prime \sin \alpha_k)^2$ increases, $ g_{k,max}(\boldsymbol{z}_k|\boldsymbol{x}_k)$ decreases monotonically. 

Let $g_{k,max}(\boldsymbol{z}_k|\boldsymbol{x}_k) = \epsilon$. Equation (\ref{Eq: likelihood_max}) can be reformed as

\begin{equation}  \label{Eq: likelihood_small_value}
\theta^\prime = \arcsin \sqrt{\frac{2\rho^2(r_k)}{r_k^2 sin^2 \alpha_k} \ln \left[ \epsilon (2\pi)^{\frac{3}{2}} \rho^3(r_k) \right]^{-1}}
\end{equation}

Since $r_k \in \left[ r_{min},\ r_{max} \right]$ and $\alpha_k \in \left[ \frac{\pi-\theta_v}{2}, \frac{\pi+\theta_v}{2} \right]$ are in close intervals, $\theta^\prime$ must have a maximum value $\theta^\prime_{max}$. For example,

\textbf{Case 1}: when $\rho(r_k)$ equals a constant value $\sigma$, $\theta^\prime_{max}$ is
\begin{equation}
  \label{Eq: max_case1}
  \theta^\prime_{max} =  \arcsin \sqrt{\frac{2\sigma^2}{r_{min}^2 cos^2  \frac{\theta_v}{2}} \ln \left[ \epsilon (2\pi)^{\frac{3}{2}} \sigma^3 \right]^{-1}}
\end{equation}

\textbf{Case 2}: when $\rho(r_k) = \sigma^\prime r_k$, which means the measurement standard deviation grows linearly with $r_k$, then
\begin{equation}\label{Eq: max_case2}
  \theta^\prime_{max} =  \arcsin \sqrt{\frac{2\sigma^{\prime 2}}{cos^2  \frac{\theta_v}{2}} \ln \left[ \epsilon (2\pi)^{\frac{3}{2}}\sigma^{\prime 3} r_{min}^3 \right]^{-1}}
\end{equation}

Therefore, given a threshold $\epsilon$, $\theta^\prime_{max}$ can be calculated and the parameter $n$ for the activation space is $n=\lceil \frac{\theta^\prime_{max}}{\theta} \rceil$. Then $\forall \boldsymbol{z}_k \notin \mathbb{A}^{\boldsymbol{x}_k}$, $g_{k}(\boldsymbol{z}_k|\boldsymbol{x}_k) \leq \epsilon \approx 0$. Note the formula in the square root symbol in Eq. (\ref{Eq: max_case1}) or (\ref{Eq: max_case2}) should be in the range $[0,1]$, which generally holds given real-world sensor parameters and robot size. Special cases when $\theta_v$ is near $\pi$ or $\rho(r_k)$ is very large can make the condition invalid. Then the strategy of increasing $r_{min}$ or decreasing $\theta_v$ in the map can be adopted to make the condition valid. The strategy increases the sphere model size or decreases the pyramid number, and thus, sacrifices part of the space to be updated.



If the measurement variances on each axis are not identical, the upper envelope of the variances can be taken as $\rho(r_k)$, and the above inference still holds. If the measurement errors on each axis are not independent, Equations (\ref{Eq: likelihood_independent}) to (\ref{Eq: max_case2}) cannot hold but the derived result can be used as an approximation to determine $n$. In the following context, we suppose the measurement errors on each axis are independent.

At the particle level, substitute $\boldsymbol{x}_k$ with $\tilde{\boldsymbol x}_k^{(i)}$, and then, $g_{k}(\boldsymbol{z}_k|\tilde{\boldsymbol x}_k^{(i)}) \approx 0$ if $\boldsymbol{z}_k \notin \mathbb{A}^{\tilde{\boldsymbol x}_k^{(i)}}$, where $\mathbb{A}^{\tilde{\boldsymbol x}_k^{(i)}}$ is the activation space of particle $\tilde{\boldsymbol x}_k^{(i)}$ and $\mathbb{A}^{\tilde{\boldsymbol x}_k^{(i)}} = \mathbb{A}^{\boldsymbol{x}_k}$ if $\tilde{\boldsymbol x}_k^{(i)}$ is in the same pyramid subspace of $\boldsymbol{x}_k$.
Let $L_k^{\mathbb{A}^{\boldsymbol{z}_k}}$ denote the number of particles whose activation space includes $\boldsymbol{z}_k$. Equations (\ref{Eq: particles_posterior_weights1}) and (\ref{Eq: particles_posterior_weights2}) can be expressed as:
\begin{align}
  \label{Eq: particles_posterior_weights1_approx}
  w_k^{(i)} =\left[ 1-P_d + \sum_{ \boldsymbol{z}_k \in \mathbb{A}^{\tilde{\boldsymbol x}_k^{(i)}}} \frac{P_d g_{k}(\boldsymbol{z}_k|\tilde{\boldsymbol x}_{k}^{(i)})}{ \kappa_k(\boldsymbol{z}_k) + \text{C}_k(\boldsymbol{z}_k) } \right]  &  w_{k|k-1}^{(i)} \\
  \label{Eq: particles_posterior_weights2_approx}
  \text{C}_k(\boldsymbol{z}_k) = \sum_{j=1}^{L_k^{\mathbb{A}^{\boldsymbol{z}_k}}} P_d w_{k|k-1}^{(j)} g_{k}(\boldsymbol{z}_k|\tilde{\boldsymbol x}_{k}^{(j)})  &
\end{align}

$L_k^{\mathbb{A}^{\boldsymbol{z}_k}}$ is about $\frac{(2n+1)^2}{N_f}$ times of $L_k$ and $N_f = \frac{\theta_h \theta_v}{\theta^2}$ is the number of pyramid subspaces in $\mathbb{M}_{f}$. Hence, the complexity of the update step in (\ref{Eq: particles_posterior_weights1_approx}) and (\ref{Eq: particles_posterior_weights2_approx}) is about $\frac{(2n+1)^2}{N_f}$ times to Equations (\ref{Eq: particles_posterior_weights1}) and (\ref{Eq: particles_posterior_weights2}).
To speed up computing, $\theta = \theta_{max}^\prime$ is adopted in practice. Then $n=\lceil \frac{\theta^\prime_{max}}{\theta} \rceil=1$ and the computational complexity is reduced to $\frac{\theta^2}{\theta_h\theta_v}$ times of (\ref{Eq: particles_posterior_weights1}) and (\ref{Eq: particles_posterior_weights2}).
Take $r_{min} = 0.15$ m, $\sigma^\prime=1\%$, and $\epsilon=0.01$ as example. With Equation (\ref{Eq: max_case2}), it can be derived that $\theta = \theta_{max}^\prime=3^\circ$ and $\frac{\theta^2}{\theta_h\theta_v} \approx 0.002$.


\subsection{Particle Birth} \label{Section: Newborn Particles}

Following the method in \cite{ImprovedSMC2010}, we generate newborn particles with measurement points $\text{Z}_k$. Since $\text{Z}_k \in \mathbb{M}_{f}$, the  newborn particles are also in $\mathbb{M}_{f}$.
For measurement point $\boldsymbol{z}_k \in \text{Z}_k$, we generate particles with a number of $L_b$.
Then the number of newborn particles in total is $M_k L_b$. The position of each newborn particle is sampled from the Gaussian noise model in Eq. (\ref{Eq: likelihood}). Normally, the velocity of the newborn particle is randomly sampled in a feasible velocity range. However, this random sampling leads to heavy noise, and the convergence speed is slow. Thus we sample the velocity of each newborn particle through an initial velocity estimation method, which is described in Section \ref{Section: Initial Velocity Estimation} and \ref{Section: Mixture Model}. The weight of these particles are set to be $\frac{v_{k|k-1}^b}{M_k L_b}$, where $ v_{k|k-1}^b = \int \gamma_{k|k-1}(\boldsymbol{x}_k) d \boldsymbol{x}_k$ is a parameter that controls the expected number of newborn objects.

According to \cite{ImprovedSMC2010}, the weight of the newborn particle is calculated separately in the update step. Then the weight update Equations (\ref{Eq: particles_posterior_weights1_approx}) and (\ref{Eq: particles_posterior_weights2_approx}) are reformed to represent the survived particles and the newborn particles separately:
\begin{align}
  \label{Eq: particles_posterior_weights1_approx_newborn}
  & w_{s,k}^{(i)} =\left[ 1-P_d + \sum_{\boldsymbol{z}_k \in \mathbb{A}^{\tilde{\boldsymbol x}_k^{(i)}}} \frac{P_d g_{k}(\boldsymbol{z}_k|\tilde{\boldsymbol x}_{k}^{(i)})}{ \kappa_k(\boldsymbol{z}_k) + \text{C}_k^{\prime}(\boldsymbol{z}_k) } \right]  w_{s,k|k-1}^{(i)}  \\
  \label{Eq: particles_posterior_weights2_approx_newborn}
  &  \qquad \qquad w_{b,k}^{(j)} = \sum_{\boldsymbol{z}_k \in \mathbb{A}^{\tilde{\boldsymbol x}_k^{(i)}}} \frac{w_{b,k|k-1}^{(j)}}{\kappa_k(\boldsymbol{z}_k) + \text{C}_k^{\prime}(\boldsymbol{z}_k)} \\
  \label{Eq: particles_posterior_weights3_approx_newborn}
  &\text{C}_k^{\prime}(\boldsymbol{z}_k) = \sum_{j=1}^{M_k L_b} w_{b,k|k-1}^{(j)}  + \sum_{j=1}^{L_{s,k}^{\mathbb{A}^{\boldsymbol{z}_k}}} P_d w_{k|k-1}^{(j)} g_{k}(\boldsymbol{z}_k|\tilde{\boldsymbol x}_{k}^{(j)})
\end{align}
where $L_{s,k}^{\mathbb{A}^{\boldsymbol{z}_k}} \leq L_{k}^{\mathbb{A}^{\boldsymbol{z}_k}}$ is the number of survived particles whose activation space includes $\boldsymbol{z}_k$. $w_{b,k|k-1}^{(j)}$ is the prior weight of the newborn particle and is $\frac{v_{k|k-1}^b}{M_k L_b}$.

\subsection{Resampling} \label{Section: Resampling}
The resampling step is to constrain the number of particles and prevent degeneration. After resampling, the cardinality expectation and the posterior PHD of $\text{X}_k$, i.e., $\mathbf{E} [|\text{X}_k|]$ and $D_{\text{X}_{k}}(\boldsymbol{x}_k)$, should not change. With $D_{\text{X}_{k}}(\boldsymbol{x}_k)$ in the form of (\ref{Eq: PHD particles_posterior}), the cardinality expectation of $\text{X}_k$ is estimated by (\ref{Eq: PHD_integral})
\begin{equation}\label{Eq: estimated_objects_num}
  \mathbf{E} [|\text{X}_k|] = \int \sum_{i=1}^{L_{k}} w_{k}^{(i)} \delta(\boldsymbol{x}_k - \tilde{\boldsymbol x}_{k}^{(i)}) \text{d}\boldsymbol{x}_k = \sum_{i=1}^{L_{k}} w_{k}^{(i)}
\end{equation}

For a single voxel subspace $\mathbb{V}_j$, the cardinality expectation $\mathbf{E} [|\text{X}_k^{(\mathbb{V}_j)}|]$ can also be estimated with the weights of the particles inside, which is
\begin{equation}\label{Eq: estimated_objects_num_voxel}
  \mathbf{E} [|\text{X}_k^{(\mathbb{V}_j)}|]= \int D_{\text{X}_k^{(\mathbb{V}_j)}}(\boldsymbol{x}) \text{d}\boldsymbol x \approx \sum_{i=1}^{L_{k}^{(\mathbb{V}_j)}} w_{k}^{(i)}
\end{equation}
where $L_{k}^{(\mathbb{V}_j)}$ represents the number of particles in $\mathbb{V}_j$ at time $k$. Although some particles outside of $\mathbb{V}_j$ but close to $\mathbb{V}_j$ may be relevant to $D_{\text{X}_k^{(\mathbb{V}_j)}}(\boldsymbol{x})$, they are not considered and thus the approximately equal sign is used.

Then the resampling is conducted by rejection sampling \cite{RejectionSampling} in each voxel subspace. 
The voxel subspace rather than the whole map is used. The reason is that if an area contains only low-weight particles, rejection sampling in the whole map may reject all these particles and decrease the occupancy probability of the area falsely.
Let $L_{max}^{\mathbb{V}}$ and $L_{max}$ denote the allowed maximum number of particles in a voxel subspace and in the map, respectively, after resampling. $L_{max} = L_{max}^{\mathbb{V}} N_v$.
Then the number of particles after resampling is
\begin{equation}\label{Eq: resampling_nums}
  \hat{L}_{k}^{\left(\mathbb{V}_{j}\right)}=\left\{\begin{array}{ll}
  L_{max}^{\mathbb{V}}, &  \ \text{if} \ L_{k}^{\left(\mathbb{V}_{j}\right)}>L_{max}^{\mathbb{V}} \\
  L_{k}^{\left(\mathbb{V}_{j}\right)}, & \text { otherwise }
  \end{array}\right.
\end{equation}

The weight of the particles in $\mathbb{V}_{j}$ after resampling is identically
\begin{equation}\label{Eq: resampling_weight}
  \hat{w}_k^{(i)} = \frac{\mathbf{E} [|\text{X}_k^{(\mathbb{V}_j)}|]}{\hat{L}_{k}^{\left(\mathbb{V}_{j}\right)}}
\end{equation}

\section{Extensions in Mapping} \label{Section: extensions in mapping}
This section proposes some important extension modules.
Firstly, the initial velocity estimation module for newborn particles and a mixture model composed of a static model and a constant velocity model are proposed to reduce the noise in mapping. Then the occupancy status estimation and future status prediction modules, which generate the output designed for motion planning, are expressed. Finally, several useful extra extensions are discussed.

\subsection{Initial Velocity Estimation} \label{Section: Initial Velocity Estimation}
\begin{figure}
\centerline{\psfig{figure=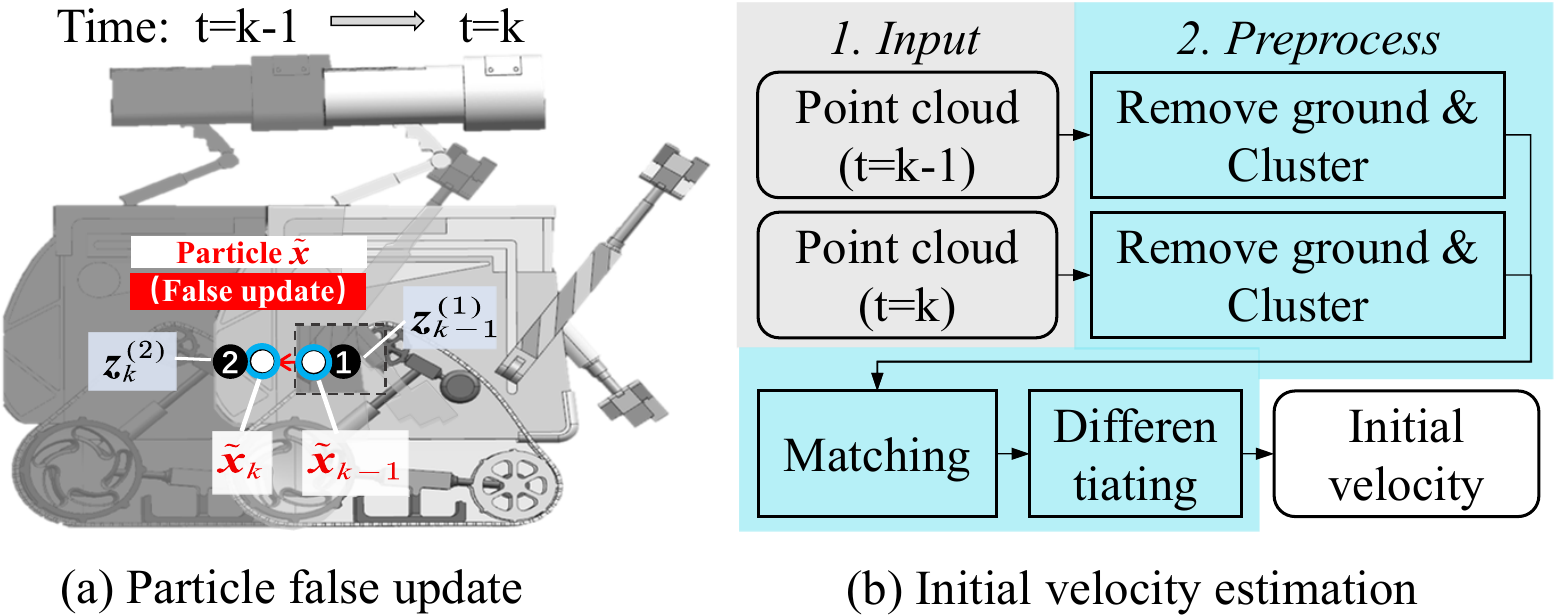, width=3.1in}}
\caption{(a) Illustration of the particle false update problem and (b) the initial velocity estimation procedures. In subfigure (a), a dynamic obstacle (a mobile robot) moves rightwards from $t=k-1$ to $t=k$. At $t=k-1$, a measurement point $\boldsymbol{z}^{(1)}_{k-1}$, on the right side of the obstacle, is observed. With $\boldsymbol{z}^{(1)}_{k-1}$, a particle with state $\tilde{\boldsymbol{x}}_{k-1}$ is generated in the particle birth step (Section \ref{Section: Newborn Particles}). Since $\boldsymbol{z}^{(1)}_{k-1}$ doesn't provide velocity, the commonly used way is to give $\tilde{\boldsymbol{x}}_{k-1}$ a random velocity, e.g., a velocity to the left. At $t=k$, following the CV model, $\tilde{\boldsymbol{x}}_{k-1}$ moves to $\tilde{\boldsymbol{x}}_{k}$. 
Since the particle's velocity is the opposite to the obstacle's velocity, this particle should belong to noise and its weight should be decreased in the update procedure at $k$. However, due to the large size of the robot, another measurement point $\boldsymbol{z}^{(2)}_{k}$, close to $\tilde{\boldsymbol{x}}_{k}$, may be observed at $k$ and falsely increased the weight of the particle in the update step. 
We call this problem the particle false update.
Note the problem also exists in modeling large-size static obstacles and is more frequent when multiple dynamic and static obstacles exist, in which case particles generated from one obstacle may be falsely updated with the measurement from another obstacle.
}
\label{Fig: velocity estimation}
\end{figure}

The particle-based maps model the obstacles as point objects. This model is very friendly with particle-based tracking but works only at the sub-object level, which will cause non-negligible noise when the obstacle has a relatively large volume.
Specifically, the noise is caused by the false update of the particles. Fig. \ref{Fig: velocity estimation} (a) illustrates the false update. The false update leads to many particles with a large weight but a wrong velocity, and further causes heavy noise in predicting the occupancy status of the area out of the FOV or at a future time. 
When the velocity of the newborn particle is randomly generated, the particle false update problem occurs frequently.

To alleviate the problem and reduce noise, we add an object-level estimation by considering initial velocities for the newborn particles. The procedures to acquire the initial velocities from two adjacent point clouds are shown in Fig. \ref{Fig: velocity estimation} (b).
The point cloud that obviously belongs to static obstacles, like the ground, is segmented by considering the height dimension and assigned zero velocity.
The rest point cloud is clustered, and the result clusters are matched with the clusters extracted from the last frame. Then the velocity of each cluster can be estimated by differentiating the position of the matched clusters' centers. We use the Euclidean cluster extraction based on k-d tree \cite{Clustering2010} for clustering and the Kuhn-Munkras (KM) algorithm for matching. In the matching process, the position of the cluster center and the number of points in the cluster are used as features. If a cluster at time $k$ cannot be matched, this cluster is regarded as a new obstacle, and no velocity estimation result is assigned.

The velocity estimated by position differentiating between two adjacent inputs is quite noisy because of three main reasons. The first reason is that the position error of the point cloud measurement is amplified and propagated to the velocity estimation by differentiating. The second reason is that the position of a cluster center varies when using point clouds observed from different angles, and the third is that the clustering and matching result might contain many errors in complex environments. We have assumed that the measurement noise of the point cloud is Gaussian noise. Thus the noise caused by the first reason is still Gaussian noise. However, the noise caused by the latter two reasons can be very random. Therefore, the estimated velocity cannot be regarded as the velocity measurement and utilized in the update step. We thus adopt this estimated velocity as a reference of initial particle velocities in the particle birth step on the basis of a mixture model. Details are presented in Section \ref{Section: Mixture Model}. 

\subsection{Mixture Model} \label{Section: Mixture Model}
To further reduce the noise caused by the false update and model static objects better, we adopt a mixture model. The mixture model supposes the state of a point object is the combination of two components, i.e., $\boldsymbol{x}_k = \lambda_1 \boldsymbol{x}_{k,d} + \lambda_2 \boldsymbol{x}_{k,s}$. 
$\lambda_1$ and $\lambda_2$ are weight coefficients that satisfy $\lambda_1 + \lambda_2 = 1$. 
$\boldsymbol{x}_{k,d}$ is a dynamic object state component. 
$\boldsymbol{x}_{k,s}$ is a static object state component with zero velocity. 
Since the environment is unknown, the value of $\lambda_1$ and  $\lambda_2$ should not be fixed but should be updated in the filtering process.
We assume that the objects in one voxel subspace, a small subspace, have the same weight coefficients. 
In each voxel subspace, the dynamic object states and the static object states can be regarded as two independent RFSs, $\text{X}_d^{(\mathbb V)}$ and $\text{X}_s^{(\mathbb V)}$, respectively.
Then $\lambda_1$ and $\lambda_2$ is estimated by the ratio between $|\text{X}_d^{(\mathbb V)}|$ and $|\text{X}_s^{(\mathbb V)}|$.

Using the property described in Equation (\ref{Eq: PHD_integral}) and the same deduction in Equation (\ref{Eq: estimated_objects_num}), $|\text{X}_d^{(\mathbb V)}|$ and $|\text{X}_s^{(\mathbb V)}|$ can be estimated with the weight summation of the particles.
A particle $\tilde{\boldsymbol x}^{(i)}$ might correspond to $\boldsymbol{x}_{k,d}$ or $\boldsymbol{x}_{k,s}$, depending on the transition density $\pi_{k|k-1}(\boldsymbol{x}_{k,d}|\tilde{\boldsymbol x}_{k-1}^{(i)})$ or $\pi_{k|k-1}(\boldsymbol{x}_{k,s}|\tilde{\boldsymbol x}_{k-1}^{(i)})$. For simplicity, the particle's speed, namely $V(\tilde{\boldsymbol x}^{(i)})$ is used as the feature to determine the correspondence, and the Dempster Shafer theory (DST) is adopted to approximate $|\text{X}_d^{(\mathbb V)}|$ and $|\text{X}_s^{(\mathbb V)}|$. The time subscript $k$ is omitted to simplify the notation.
The universe of DST, in our case, is $U = \left\{ \text{d}, \text{s} \right\}$, where $\text{d}$ is the dynamic hypothesis and $\text{s}$ is the static hypothesis. The power set is $2^U = \left\{\emptyset,  \left\{\text{d}\right\}, \left\{\text{s}\right\}, U \right\}$. The mass function $m(A)$ has the properties that $\sum_{A \in 2^{U}} m(A)=1$  and $m(\emptyset) = 0$.

We suppose the weight summation of particles that satisfy $V(\tilde{\boldsymbol x}^{(i)}) = 0$ is $W_s^{(\mathbb V)}$ and the weight summation of particles that satisfy $V(\tilde{\boldsymbol x}^{(i)}) \geq \hat{V}$ is $W_d^{(\mathbb V)}$, where $\hat{V}$ is a threshold suggesting that particles with a velocity larger than $\hat{V}$ correspond to $\boldsymbol{x}_{k,d}$ rather than $\boldsymbol{x}_{k,s}$. The particles with $ 0 < V(\tilde{\boldsymbol x}^{(i)}) < \hat{V}$, however, can correspond to $\boldsymbol{x}_{k,d}$ or $\boldsymbol{x}_{k,s}$. Suppose the weight summation of these particles is $W_{d,s}^{(\mathbb V)}$. Then the masses are defined with
\begin{equation}\label{Eq: masses}
\begin{aligned}
  & m(\left\{\text{d}\right\}) = \frac{W_d^{(\mathbb V)}}{W^{(\mathbb V)}}, \  \  m(\left\{\text{s}\right\}) = \frac{W_s^{(\mathbb V)}}{W^{(\mathbb V)}} \\
  & m(U) = \frac{W_{d,s}^{(\mathbb V)}}{W^{(\mathbb V)}},  \  \ W^{(\mathbb V)}=W_d^{(\mathbb V)} + W_s^{(\mathbb V)}+W_{d,s}^{(\mathbb V)}
\end{aligned}
\end{equation}
which describes the basic belief.
Then the belief and the plausibility are
\begin{equation}\label{Eq: belief1}
\begin{aligned}
& bel(\left\{\text{d}\right\}) = m(\left\{\text{d}\right\}), \ pl(\left\{\text{d}\right\}) = m(\left\{\text{d}\right\}) +  m(U) \\
& bel(\left\{\text{s}\right\}) = m(\left\{\text{s}\right\}), \ pl(\left\{\text{s}\right\}) = m(\left\{\text{s}\right\}) +  m(U)
\end{aligned}
\end{equation}

According to DST, the probability is between the belief and the plausibility. We simply take the median as the probability estimation, which is $pr(\cdot)=\frac{bel(\cdot) + pl(\cdot)}{2}$. The cardinalities of dynamic and static objects in a voxel are approximated by
\begin{equation}\label{Eq: DST cardinality estimation}
\begin{aligned}
  & |\text{X}_d^{(\mathbb V)}| \approx W^{(\mathbb V)} \left[ bel(\left\{\text{d}\right\}) + pl(\left\{\text{d}\right\}) \right]  / 2 \\
  & |\text{X}_s^{(\mathbb V)}| \approx W^{(\mathbb V)} \left[ bel(\left\{\text{s}\right\}) + pl(\left\{\text{s}\right\}) \right]  / 2
\end{aligned}
\end{equation}

Then the coefficients $\lambda_1$ and $\lambda_2$ are estimated with the ratio of the cardinalities:
\begin{equation}\label{Eq: lambda estimation}
  \small
  \begin{aligned}
  &\lambda_1 = \frac{|\text{X}_d^{(\mathbb V)}|}{|\text{X}_d^{(\mathbb V)}| + |\text{X}_s^{(\mathbb V)}|} = \frac{W_d^{(\mathbb V)}}{W^{(\mathbb V)}}+\frac{1}{2} \frac{W_{d, s}^{(V)}}{W^{(\mathbb V)}} \\
  &\lambda_2 = \frac{|\text{X}_s^{(\mathbb V)}|}{|\text{X}_d^{(\mathbb V)}| + |\text{X}_s^{(\mathbb V)}|} = \frac{W_s^{(\mathbb V)}}{W^{(\mathbb V)}}+\frac{1}{2} \frac{W_{d, s}^{(V)}}{W^{(\mathbb V)}}
  \end{aligned}
\end{equation}

In the prediction step, the motion model in Eq. (\ref{Eq: single_object_motion}) from time $k-1$ to $k$ turns to
\begin{equation}\label{Eq: mixture model sample}
  \boldsymbol{x}_k=\lambda_1\left[f_Q\left(\boldsymbol x_{k-1, d}\right)+\boldsymbol\Sigma\right]+\lambda_2\left[\boldsymbol x_{k-1, s}+\boldsymbol\Sigma^{\prime}\right]
\end{equation}
where $\boldsymbol\Sigma^{\prime}$ is the Gaussian noise whose velocity dimension is zero. At the particle level, with Eq. (\ref{Eq: lambda estimation}) it can be derived that particles satisfying $V(\tilde{\boldsymbol x}^{(i)}) > \hat{V}$ and half number of particles satisfying $0 < V(\tilde{\boldsymbol x}^{(i)}) < \hat{V}$ correspond to $\boldsymbol x_{k-1, d}$, and their prior states are sampled with Eq. (\ref{Eq: particle prediction cv}). The rest particles correspond to $\boldsymbol x_{k-1, s}$, and their prior states are sampled with $\tilde{\boldsymbol{x}}_{k \mid k-1}^{(i)}=\tilde{\boldsymbol{x}}_{k-1}^{(i)}+\boldsymbol{u}^\prime$, where $\boldsymbol{u}^\prime$ doesn't contain velocity noise. The update step remains the same because the measurement does not contain velocity observation.

In the particle birth step, the velocities of the newborn particles are assigned based on $\lambda_1$ and $\lambda_2$ in the corresponding voxel subspace, and the initial velocity estimation results in Section \ref{Section: Initial Velocity Estimation}. If a measurement point is labeled static, e.g., the ground, in the initial velocity estimation procedure, the velocities of the particles generated from this point are all zero. Otherwise, the mixture model is used. The number of dynamic particles generated from a measurement point is $\lambda_1 L_b$, and the number of static particles is $\lambda_2 L_b$. 
According to the discussion of estimation noise in Section \ref{Section: Initial Velocity Estimation}, the velocities of dynamic particles are composed of two parts: velocities sampled from a Gaussian distribution and random velocities. Since real-world sensors usually contain heavy noise, we set a large variance for the Gaussian distribution, and the particles with random velocities take $0.5 \lambda_1L_b$.
If too few particles exist in the voxel subspace where the measurement point belongs, e.g., the situation when the voxel subspace is observed for the first time, an initial guess of $\lambda_1 = \lambda_2 = 0.5$ is used.


\subsection{Occupancy Status Estimation} \label{Section: Occupancy Status Estimation}  
\begin{figure*}
\centerline{\psfig{figure=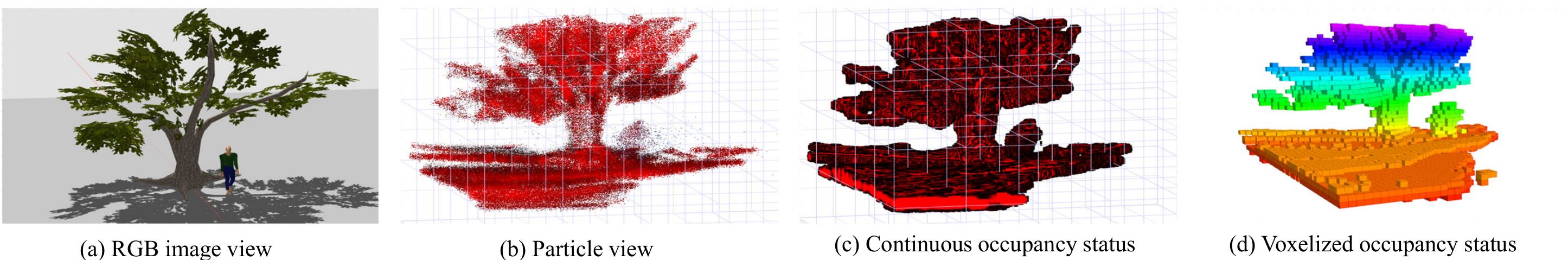,width=6.8in}}
\caption{Occupancy status estimation of the DSP map in a scenario with a static obstacle and a dynamic obstacle. (a) shows the scenario with an RGB image. (b) illustrates the particles in the current DSP map. The color is painted in the HSV color space, with the hue keeping red and the saturation indicating the particle's weight. Higher saturation, i.e., a greater proportion of red relative to black, indicates a larger weight. The estimated occupancy status in a continuous form is shown in (c). Higher saturation indicates a larger occupied probability. (d) utilizes the voxel subspaces to calculate a 3d grid map. The color of the grids changes with their $z$-axis height. The obstacles are from a scenario in the pedestrian street world in Fig. \ref{Fig: testing scenarios}, and the map is built with the recorded flight data used in Section \ref{Section: Mapping performance comparison}. Some parts of the tree and the ground are missing because they haven't been observed.}
\label{Fig: occupancy}
\end{figure*}

At an arbitrary point $\boldsymbol{p}$ in the map, the occupancy status is estimated by the cardinality expectation of point objects in a small neighborhood space of $\boldsymbol{p}$. Assume the point objects representing an obstacle are uniformly distributed in the space occupied by the obstacle and has no overlap. The distance between two adjacent point objects is $l^\prime$. Then, in a cubic neighborhood space with side length $l^\prime$ and centered by $\boldsymbol{p}$, there should be either one or no point object. In our case, the point cloud is pre-filtered by a voxel filter with resolution $Res$. Thus, $l^\prime = Res$.
Denote the cubic neighborhood space by $\mathbb{V}_p$ and the RSF composed of the point objects in $\mathbb{V}_p$ by $\text{X}_k^{\mathbb{V}_p}$. According to Equation (\ref{Eq: PHD_integral}) and (\ref{Eq: PHD particles_posterior}), the expectation of the cardinality of $\text{X}_k^{\mathbb{V}_p}$ is calculated with
\begin{equation} \label{Eq: occupancy status esimtaion}
  \mathbf{E} [|\text{X}_k^{\mathbb{V}_p}|] = \int D_{\text{X}_k^{\mathbb{V}_p}}(\boldsymbol{x}) \text{d}\boldsymbol x \approx \sum_{\tilde{\boldsymbol x}_{k}^{(i)} \in \mathbb{V}_p} w_k^{(i)}
\end{equation}
which is the weight summation of particles in $\mathbb{V}_p$. 

We denote the occupancy probability at $\boldsymbol{p}$ by $P_{occ}(\boldsymbol{p})$. Since $\mathbf{E}[|\text{X}_k^{\mathbb{V}_p}|]$ represents the expectation of the point object number in $\mathbb{V}_p$, and $\mathbb{V}_p$ is occupied as long as there is a point object inside, $P_{occ}(\boldsymbol{p})$ can be estimated by $P_{occ}(\boldsymbol{p}) = \mathbf{E} [|\text{X}_k^{\mathbb{V}_p}|]$, if $\mathbf{E} [|\text{X}_k^{\mathbb{V}_p}|] \leq 1$. 
In practice, $\mathbf{E} [|\text{X}_k^{\mathbb{V}_p}|]$ can be larger than one because of the noise in the input data and 
camera motions.
If $\mathbf{E} [|\text{X}_k^{\mathbb{V}_p}|] > 1$, $P_{occ}(\boldsymbol{p}) = 1$ is adopted.

The occupancy probability of a general voxel subspace, such as the voxel subspace $\mathbb{V}_{i}$ defined in Section \ref{Section: world model} with side length $l$, is estimated with
\begin{equation}\label{Eq: occupancy status esimtaion voxel}
  P_{occ}(\mathbb{V}_{i})=\left\{\begin{array}{ll}
    \text{Min}\{ \mathbf{E} [|\text{X}_k^{(\mathbb V_i)}|] \cdot (\frac{l^\prime}{l})^3, 1\}, \ \text{if} \ l \leq l^\prime \\
    \text{Min}\{ \mathbf{E} [|\text{X}_k^{(\mathbb V_i)}|], 1\}, \ \text{otherwise }
  \end{array}\right.
\end{equation}
where $\mathbf{E} [|\text{X}_k^{(\mathbb V_i)}|]$ is calculated with weight summation of particles in $\mathbb{V}_{i}$ like Equation (\ref{Eq: occupancy status esimtaion}). A scale factor $(\frac{l^\prime}{l})^3$ is applied because the volume of $\mathbb{V}_{i}$ is $(\frac{l}{l^\prime})^3$ times smaller than $\mathbb{V}_p$. If $l > l^\prime$, the estimated point object number $\mathbf{E} [|\text{X}_k^{(\mathbb V_i)}|]$ can be larger than one even if the estimation has no error. If $\mathbf{E} [|\text{X}_k^{(\mathbb V_i)}|] > 1$, $P_{occ}(\mathbb{V}_{i}) = 1$ is adopted.
With the occupancy probability, a probability threshold can then be used to get a binary occupancy status, i.e., occupied or free.
Fig. \ref{Fig: occupancy} shows an example occupancy estimation result in a scenario with a static obstacle and a dynamic obstacle. The voxelized map is shown in Subfigure (d).
It should be noted that this voxelized map doesn't suffer from the grid size problem because the mapping process is realized in the continuous space. 


\subsection{Future Occupancy Status Prediction} \label{Section: Future Occupancy Status Prediction}
\begin{figure}
  \centerline{\psfig{figure=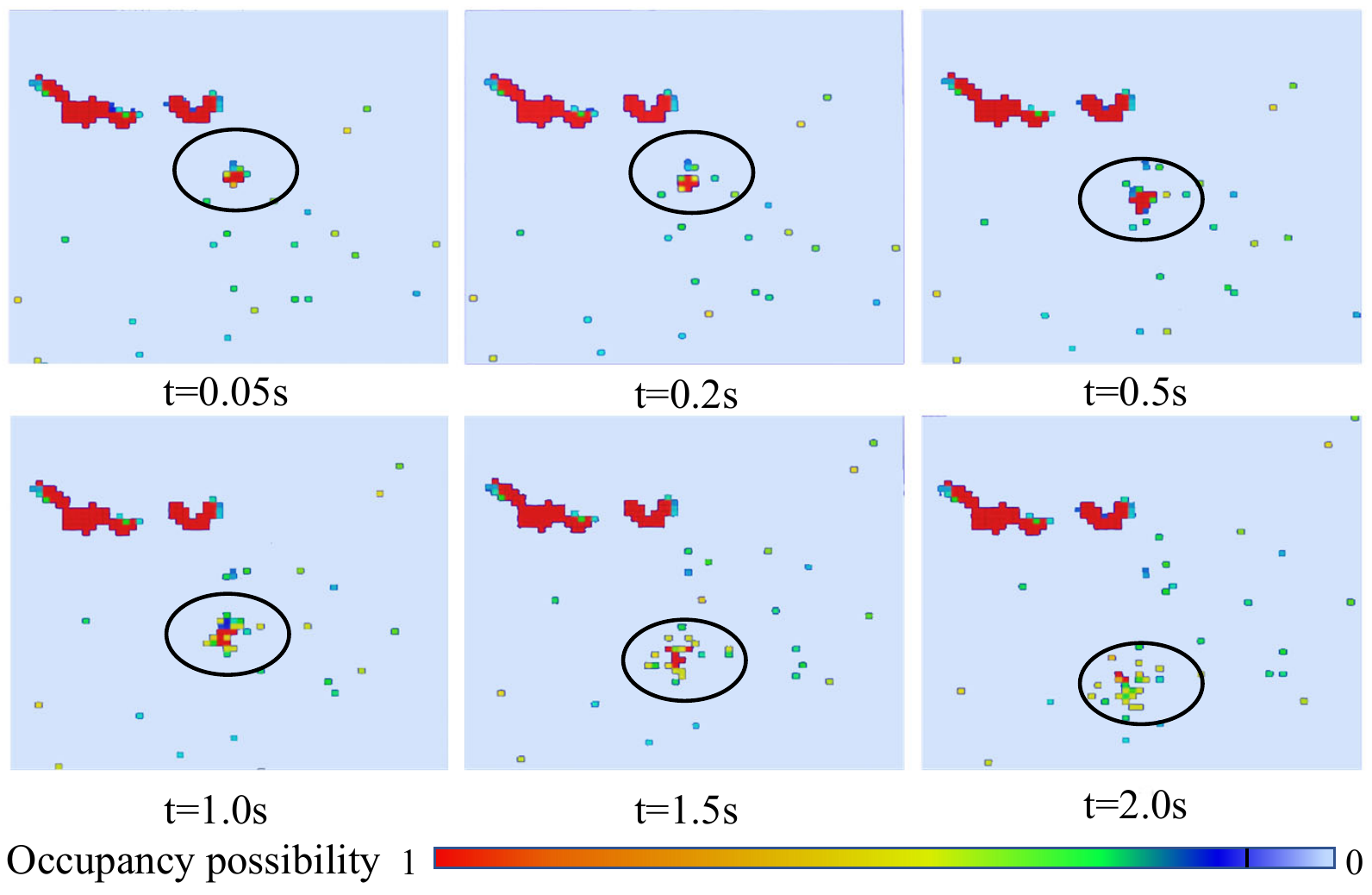, width=3.3in, height=2.0in}}
  \caption{Future occupancy status prediction. We predict the future occupancy status of the scenario in Fig. \ref{Fig: occupancy} at six future times. Only the layer $z=1.6m$ is shown to have a clear view. The black ellipses show the predicted occupancy status of the pedestrian. The pedestrian walks with a constant velocity to the bottom side. The red occupied area in the upper left corresponds to the tree's trunk and branches.}
  \label{Fig: future_status}
\end{figure}

Predicting the future occupancy status is very useful for motion planning in dynamic environments. In our DSP map, the future occupancy status prediction is fulfilled by predicting the position of the particles according to the motion model in (\ref{Eq: single_object_motion}) and (\ref{Eq: mixture model sample}), and then using (\ref{Eq: occupancy status esimtaion}) and (\ref{Eq: occupancy status esimtaion voxel}) for occupancy status estimation. Fig. \ref{Fig: future_status} presents the future occupancy estimation results of the map shown in Fig. \ref{Fig: occupancy}. The occupancy status of the static obstacle, the tree, almost stays the same in each plot. The occupied position of the dynamic obstacle, the pedestrian, is predicted to move down with the CV model. The occupied grids are spreading, and their occupancy probabilities are getting lower as the prediction time increases. The reason is the uncertainty in velocity estimation, which is reflected by the variance of particles' velocities. The estimation uncertainty also causes noise in other parts of the plots. 
Since future occupancy status prediction in dynamic environments has inevitable uncertainty, the predicted occupancy probability can be used as the risk in motion planning algorithms.



\subsection{DSP Static Map} \label{Section: DSP Static Map}
By assuming the point objects as static objects and using only the static model described in Section \ref{Section: Mixture Model},  the DSP map turns to a static map, named the DSP-Static map. In this case, the number of particles used in this map can be very small since the velocity dimension is not considered, which means the DSP-Static map is more computationally efficient. Compared to the voxel map for static environments, the DSP-Static map is continuous and free from the voxel size problem. In the experiment section, the DSP-Static map is also tested. 

\section{Implementation} \label{Section: Implementation}
This section describes an implementation of the DSP map. The implementation includes the data structure to realize subspace division and particle storage, and the specific algorithms used to build the DSP map.

\subsection{Data Structure}
The number of particles in the map can be up to one million. Thus, storage and operation of the particles are important to efficiency. Three techniques are utilized to improve efficiency: 1) The voxel subspaces are used to store particles while the pyramid subspaces only store the indexes of particles; 2) Large arrays with preallocated size rather than unordered sets, which represent RFSs natively, are used to store elements in an RFS; 3) The operations of adding and deleting particles are simplified using a flag variable. 
The first technique is to reduce memory consumption, while the second is to avoid dynamic memory allocation and increase the cache hit rate. The last technique is employed to simplify operations on particles. Detailed data structure can be found in Appendix \ref{Appendix: Algorithms for Data Structure}.

\subsection{Mapping Algorithms} \label{Section: Mapping Algorithms}
\begin{figure}
  \centerline{\psfig{figure=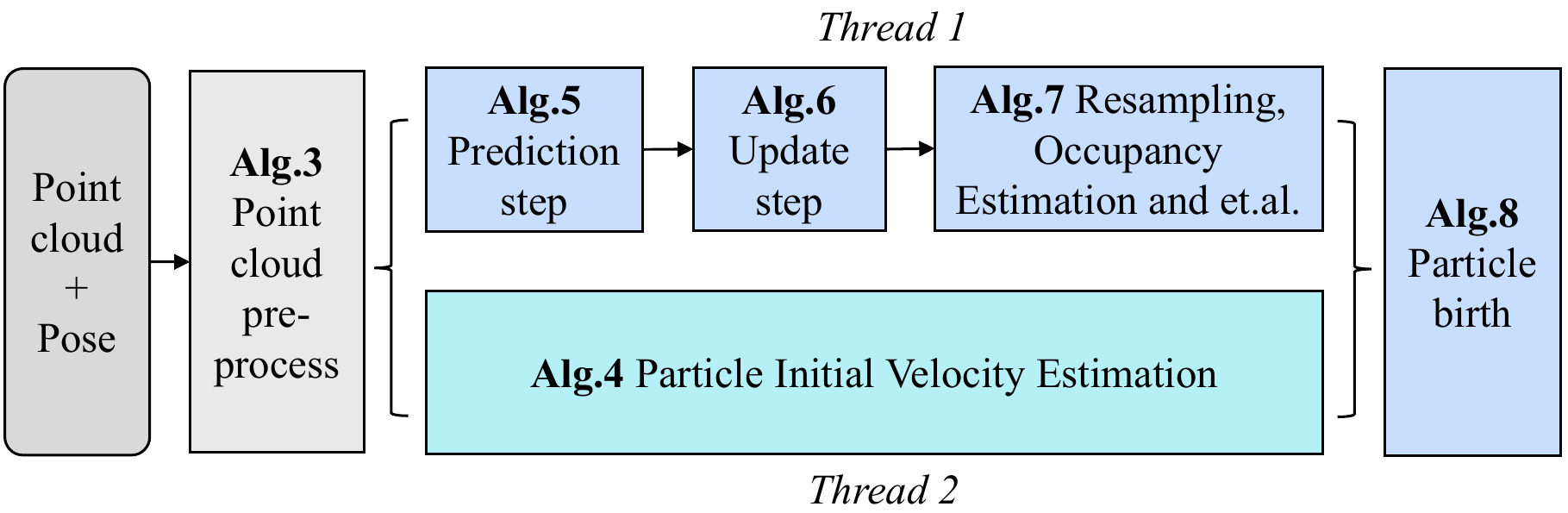, width=3.5in}}
  \caption{Flowchart of the algorithms used to implement the DSP map.}
  \label{Fig: algorithm_flowchart}
\end{figure}

A flowchart showing the order of the algorithms used for mapping is presented in Fig. \ref{Fig: algorithm_flowchart}. After the input point cloud is pre-filtered by a voxel filter with resolution $Res$ and transformed to the map frame,
two threads are opened to run the particle initial velocity estimation in parallel with prediction, update and resampling. Resampling, occupancy estimation and mixture model coefficients calculation are conducted in one loop to improve efficiency. Future occupancy status prediction is not shown but is realized by predicting particle states to more future times in the prediction step. Detailed algorithms can be found in Appendix \ref{Appendix: Algorithms for Mapping}.

\section{Experimental Results} \label{Section: experiments}
This section first evaluates the velocity estimation precision of the DSP map since velocity estimation ability is a major difference between static maps and dynamic maps.
Then the DSP map is tested with different parameters to evaluate the effect of the parameters on mapping performance and identify the best parameter values. With the identified parameter values, the mapping performance is further compared with existing works in different simulation worlds with different resolutions.  
To test the practicality of using the DSP map on robotics platforms, we also show computational efficiency comparison results on an NVIDIA Jetson board. Finally, a demo of using this map for drone obstacle avoidance is presented.

\subsection{Velocity Estimation} \label{Section: Velocity Estimation Results}
\begin{figure}
  \centerline{\psfig{figure=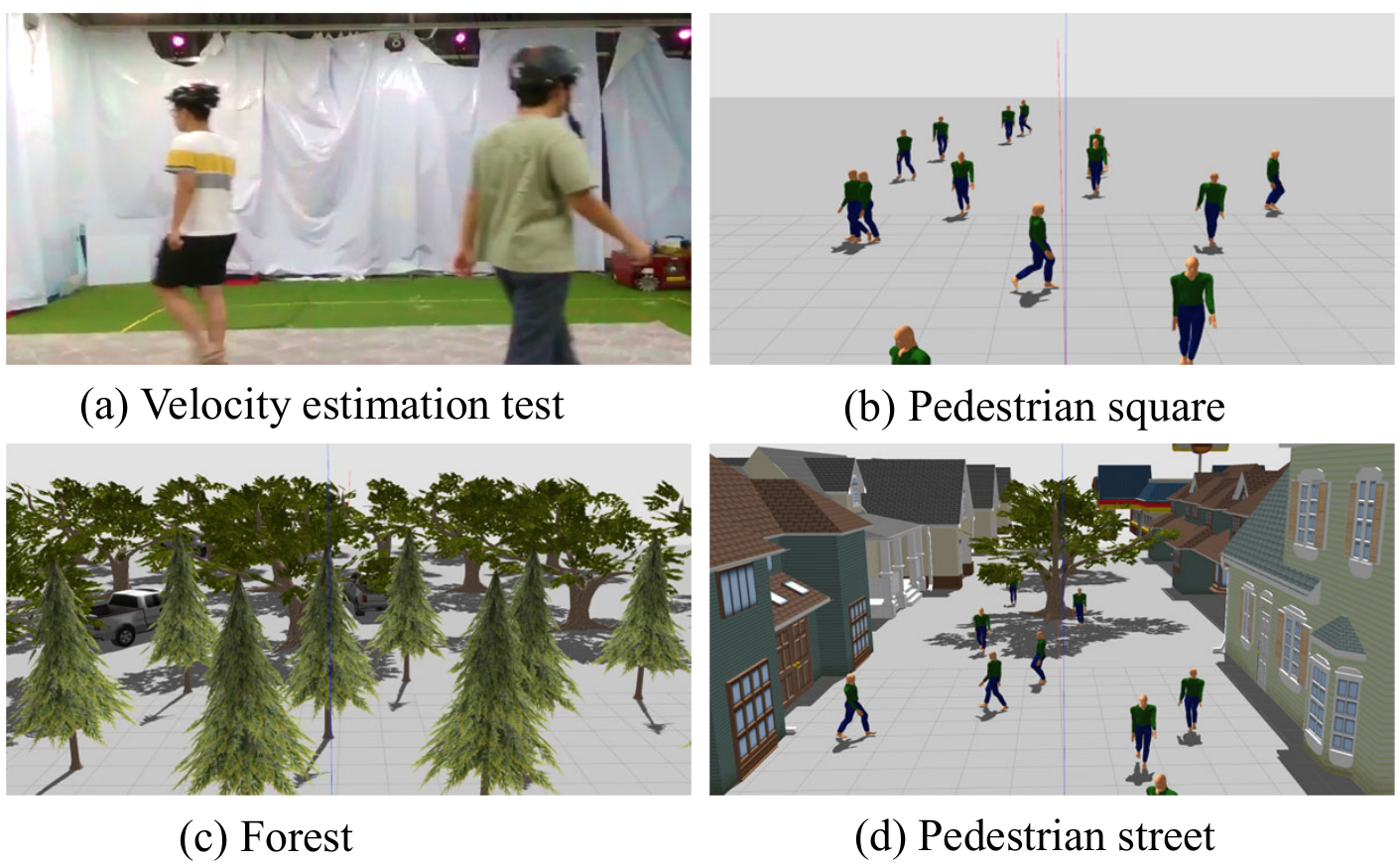, width=3.1in}}
  \caption{Real-world scenario for velocity estimation test (a), and simulation worlds for mapping performance comparison (b)-(d). The pedestrian square world (b) contains only dynamic obstacles (the ground is excluded in the evaluation tests), while the forest world (c) contains only static obstacles. The pedestrian street world (d) contains both static obstacles and dynamic obstacles.}
  \label{Fig: testing scenarios}
\end{figure}

The velocity estimation experiments were conducted with the data collected in an indoor testing field with the Nokov motion capture system. An Intel Realsense d435 camera was fixed at an edge of the testing field to collect the point cloud. Two pedestrians, wearing helmets with markers, walked around in the testing field, and their trajectories estimated by the motion capture system were recorded synchronously with the point cloud. The experiments can be divided into two groups. In the first group, the pedestrians tried to walk at a constant velocity. In the second group, the pedestrians walked randomly and freely.
Fig. \ref{Fig: testing scenarios} (a) shows the data collection scenario.

\begin{figure}[h]
  \centerline{\psfig{figure=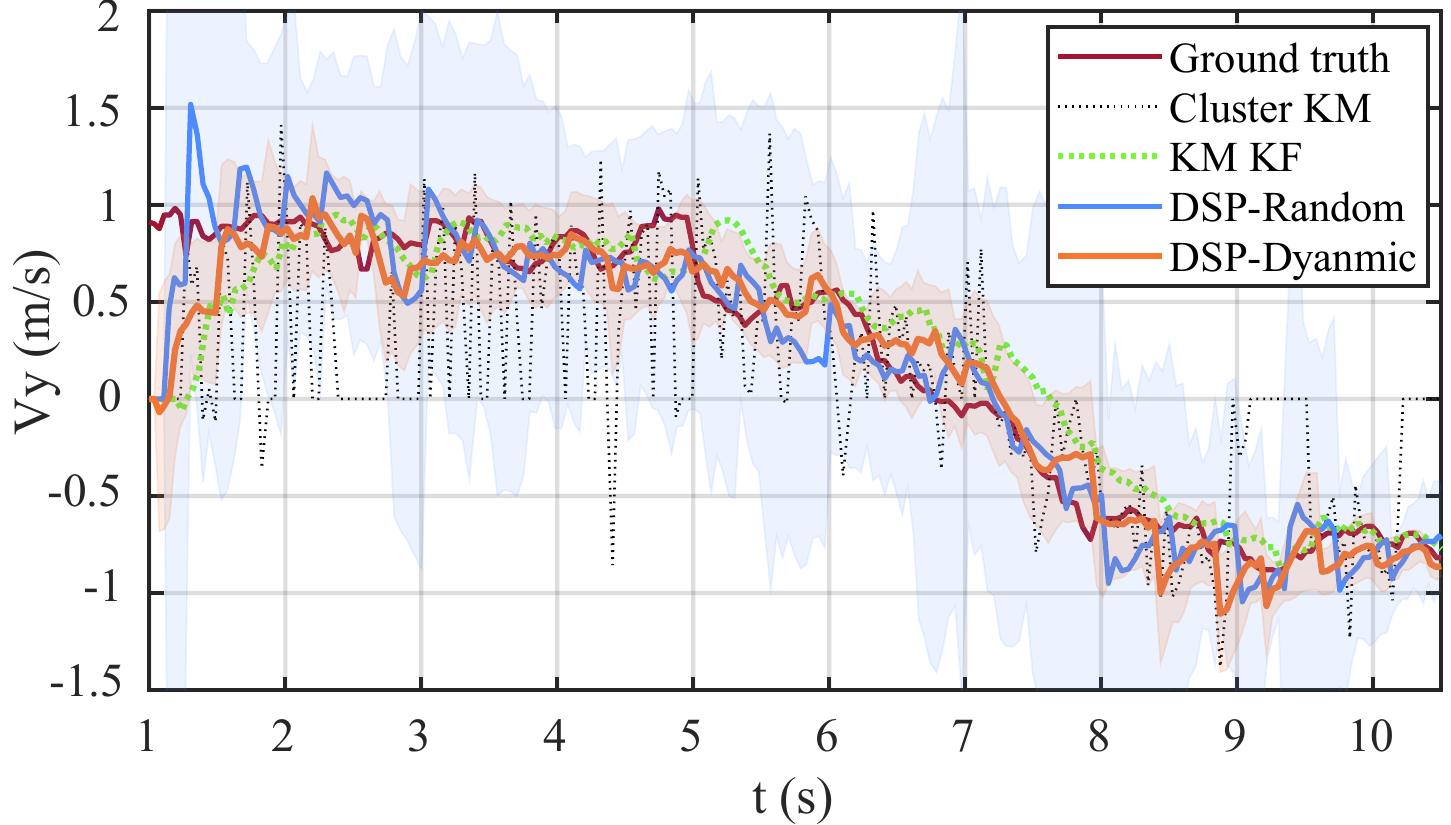, width=3.3in}}
  \caption{Velocity estimation curves of a typical tracklet. The serrated orange and blue background show the variance of the estimation results from the DSP-Dynamic map and the DSP-Random map, respectively. At $t=6s$ to $t=8s$, the pedestrian turns back.} 
  \label{Fig: velocity estimation curve 1}
\end{figure}

We compared the velocities estimated by four different point-cloud-based methods. The first method differentiates the center position of two matched clusters, and no filter is adopted. The matching is achieved by the KM algorithm. The second is a multi-object tracker realized by the KM algorithm and Kalman Filters (KF) with a CV model. The input of the KF is the center positions of matched clusters.
The third method is the DSP map with the suffix ``Random'',  whose newborn particles have random velocities. The fourth method is the DSP map with the suffix ``Dynamic'', whose newborn particles consider initial velocity estimation. Since our maps do not explicitly segment the objects, the state of a pedestrian was estimated with the particle cluster near the pedestrian's real position. Table \ref{Table: Velocity estimation result} presents the estimation results of the two groups. We consider a pedestrian walking from one side of the testing field to another a tracklet. Over thirty tracklets were collected in each group. Fig. \ref{Fig: velocity estimation curve 1} shows the velocity estimation curves of a typical tracklet.  

\begin{table}[]
\caption{Velocity estimation results of different methods.}
\centering
\begin{tabular}{l|lll|lll}
\hline
Group       & \multicolumn{3}{l|}{Constant velocity}           & \multicolumn{3}{l}{Random walking}               \\ \hline
Metric      & RMSE           & Var.           & MBD            & RMSE           & Var.           & MBD            \\ \hline
KM-Diff.    & 0.583          & -              & -              & 0.656          & -              & -              \\
KM-KF       & 0.286          & 0.479          & 0.470          & 0.309          & 0.481          & 0.476          \\
DSP-Random  & 0.332          & 1.083          & 0.641          & 0.353          & 1.077          & 0.641          \\
DSP-Dynamic & \textbf{0.277} & \textbf{0.318} & \textbf{0.398} & \textbf{0.302} & \textbf{0.335} & \textbf{0.417} \\ \hline
\end{tabular}
\label{Table: Velocity estimation result}
\end{table}

Three metrics are used for evaluation. The root mean square error (RMSE) reflects the estimation precision evaluated with the mean of the velocity estimation distribution and the ground truth from the motion capture system. The Var. is the mean variance of the different axes on every point. The differentiating method outputs a single value rather than a distribution, and thus a dash is placed in its Var. in Table \ref{Table: Velocity estimation result}. For a particle-based map, a large variance means the particles would disperse to a large scale of the area and cause much noise in the map. Mean Bhattacharyya distance (MBD) measures the similarity between the estimated and ground-truth velocity distribution. MBD considers both mean value and variance and is a composite metric. The results in Table \ref{Table: Velocity estimation result} show that DSP-Dynamic performs best with all three metrics. The differentiated velocity has a large RMSE, and the error can be huge sometimes, as Fig. \ref{Fig: velocity estimation curve 1} shows. Using KF can reduce the error, but the Var. is over 30\% larger than that of DSP-Dynamic, and the MBD is over 12\% larger.
Compared to DSP-random, DSP dynamic decreases over 14\% on RMSE, over 68\% on Var., and over 34\% on MBD, showing the importance of the initial velocity estimation.

\subsection{Mapping with Different Parameters} \label{Section: Mapping with Different Parameters}


Inspired by \cite{RFSMap} \cite{DynamicMapICRA2021} \cite{DynamicHilbert}, we evaluated the occupancy mapping performance by assessing the binary classification results, i.e., free or occupied, of the voxel subspaces. The metrics include average precision, recall, F1-Score\footnote{F1-Score $= \frac{2 precision \cdot recall}{precision + recall}$ is a balanced metric of precision and recall.}, and time consumption of a complete mapping process.
The tested parameters include maximum particle number $L_{max}$, the voxel size $Res$ of the voxel filter for the point cloud pre-process before mapping, the pyramid subspace angle $\theta$, and the voxel subspace size $l$. When $Res$ is larger, the measurement point number $M_k$ is smaller. Each parameter was tested with three levels. A full factorial experiment was conducted with data collected in the pedestrian street world (Fig. \ref{Fig: testing scenarios} (d)), where both static and dynamic obstacles exist. The world is built in the Gazebo\footnote{Gazebo simulation software: \url{https://gazebosim.org/home}} simulation software. A simulated IRIS quadrotor with a Realsense camera is controlled manually to collect point cloud and pose data for mapping. 



To generate the ground truth occupancy map, we densely and uniformly sampled points from the mesh surfaces of the static objects in the world and generated a Euclidean Distance Field (EDF) for the objects using the sampled points. The EDF changes caused by pedestrians are updated online at each evaluation step using the mesh and pose of the pedestrians. A voxel subspace was considered occupied if the distance value at the voxel's center position was no larger than $\frac{l}{2}$ and free otherwise. We also added a label array to distinguish the observed and unobserved labels of the voxels. The array is updated using the ray-casting approach with dense rays. Only the observed voxels were considered in the evaluation.

\begin{figure*}
  \centerline{\psfig{figure=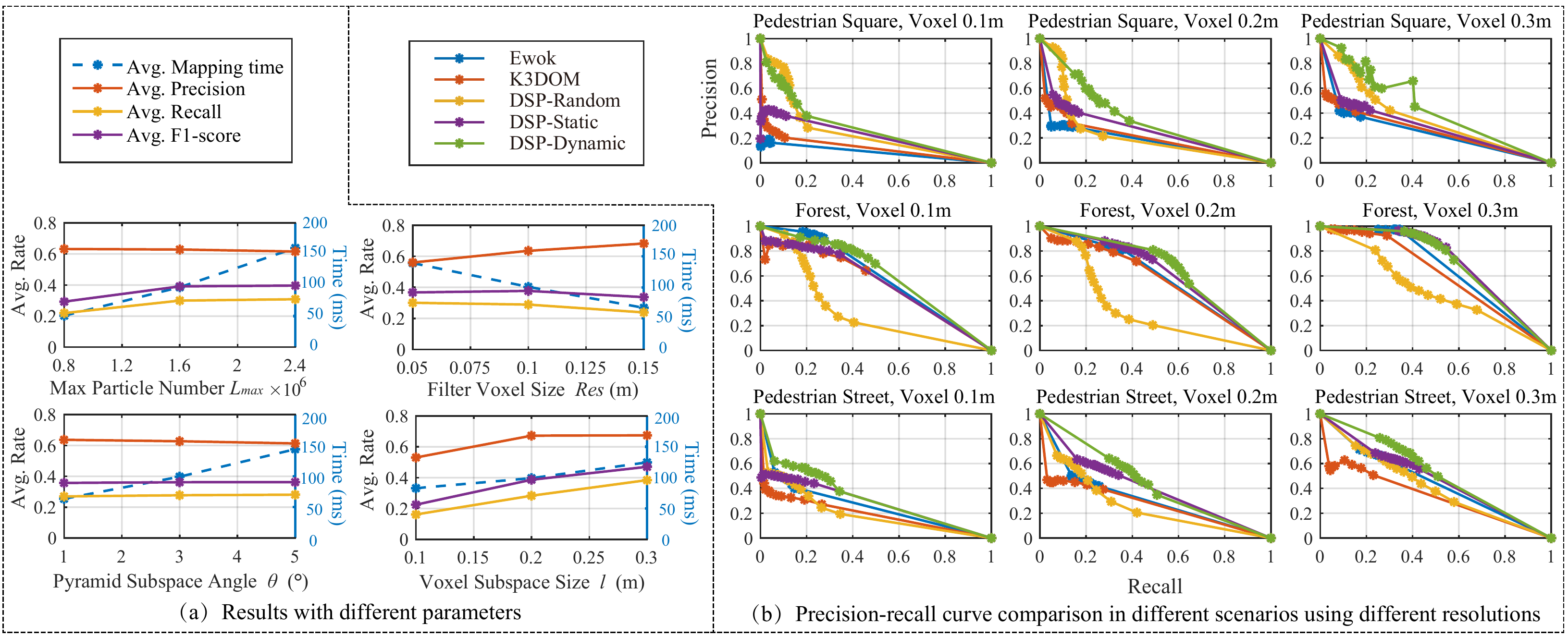, width=7.2in}}
  \caption{(a) Mapping performance with different parameters. Each parameter has three levels and is analyzed in an individual plot. (b) Precision-recall curve comparison. Results in different worlds are shown in different rows, and results with different voxel sizes are shown in different columns.}
  \label{Fig: accuracy_recall_curve_large}
\end{figure*}

The result is shown in Fig. \ref{Fig: accuracy_recall_curve_large} (a).
When $L_{max}$ increases from 0.8 million to 2.4 million, the precision does not have a noticeable change, while the average time consumption increases from 50 ms to 160 ms.
The recall rises from 0.22 to 0.30 when $L_{max}$ increases to 1.6 million but almost remains unchanged when $L_{max}$ increases further. The F1-Score has the same trend as the recall. 
Raising $Res$ leads to fewer measurement points in the point cloud and shows a positive effect on precision but a negative effect on recall. The balanced metric F1-Score reaches the maximum value of 0.38 when $Res$ is 0.1 m. The time consumption decreases as $Res$ increases. 

The pyramid subspace angle $\theta$ slightly affects precision, recall, and F1-Score. 
The F1-Score increases merely 0.005 when $\theta$ grows from one degree to five degrees. Meanwhile, the time consumption increases from 67 ms to 144 ms. 
The voxel subspace size $l$ positively correlates to all the metrics. 
When $l$ is larger, the number of voxels to classify is less, and the occupancy status of a voxel is easier to determine because more measurement points and particles are contained in one voxel. As a result, the precision, recall, and F1-Score all improve. However, a larger voxel size is usually unfavorable in motion planning. The time consumption rises because the particle operations in Algorithm \ref{Algorithm: Voxel Particle} are slower with more particles in one voxel subspace.

To achieve the best F1-Score with an acceptable time consumption (about 100 ms), $L_{max} = 1.6 \times 10^6$, $Res = 0.1$, and $\theta = 3^\circ$ are chosen. 
We further compare the performance of our map with other maps using different resolutions in the following experiment.

\subsection{Mapping Performance Comparison} \label{Section: Mapping performance comparison}

In this experiment, we compared our DSP map with a static local occupancy map named Ewok \cite{Ringbuffer} and a state-of-the-art particle-based dynamic occupancy map named K3DOM \cite{DynamicMapICRA2021}. K3DOM is the only 3D dynamic occupancy map with a released code currently.
We also compared our map with two variants: one uses newborn particles with random velocities and considers the constant velocity model only, i.e., extensions in Section \ref{Section: Initial Velocity Estimation} and \ref{Section: Mixture Model} are not adopted; another uses static newborn particles and considers the static motion model, i.e., the extension in Section \ref{Section: DSP Static Map}. To distinguish the variants, we call our map with particle initial velocity estimation and mixture model DSP-Dynamic map, and the variants DSP-Random map and DSP-static map, respectively.

\begin{figure*}[h]
  \centerline{\psfig{figure=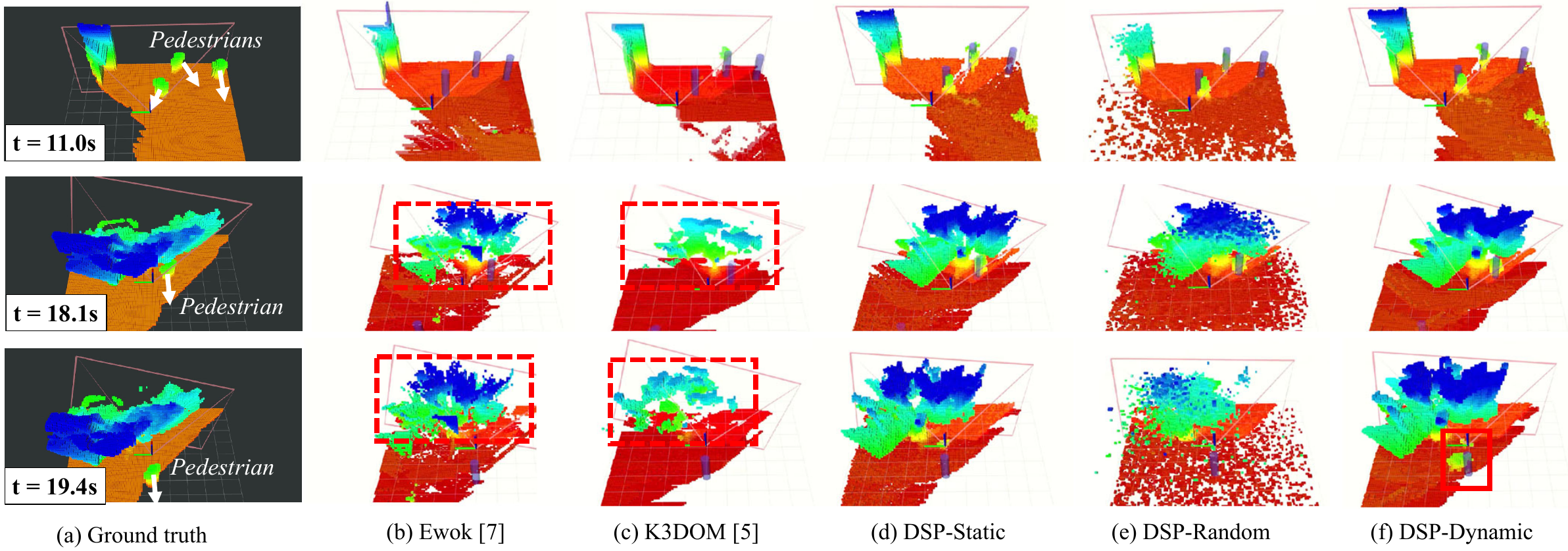,width=7.1in}}
  \caption{Snapshots of different maps in the pedestrian street world when the mapping resolution is 0.1 m. A pedestrian moves along the direction indicated by the white arrow from $t=18.1$s to $t=19.4$s.
  The color of the voxels changes with their $z$-axis height. The pink lines show the current FOV of the camera. The voxels in the FOV are painted brighter than those out of the FOV. The semi-transparent blue cylinders in the maps present the real position of the pedestrians. The red rectangle in the DSP-Dynamic map at $t=19.4$s outlines an area out of the FOV corresponding to a pedestrian. The pedestrian is out of the FOV, and thus, its occupancy status is predicted. Red dashed boxes show typical gaps and inconsistencies in grid maps when the resolution is high.}
  \label{Fig: mapping_performance_high_resolution}
\end{figure*}

\begin{figure*}[h]
  \centerline{\psfig{figure=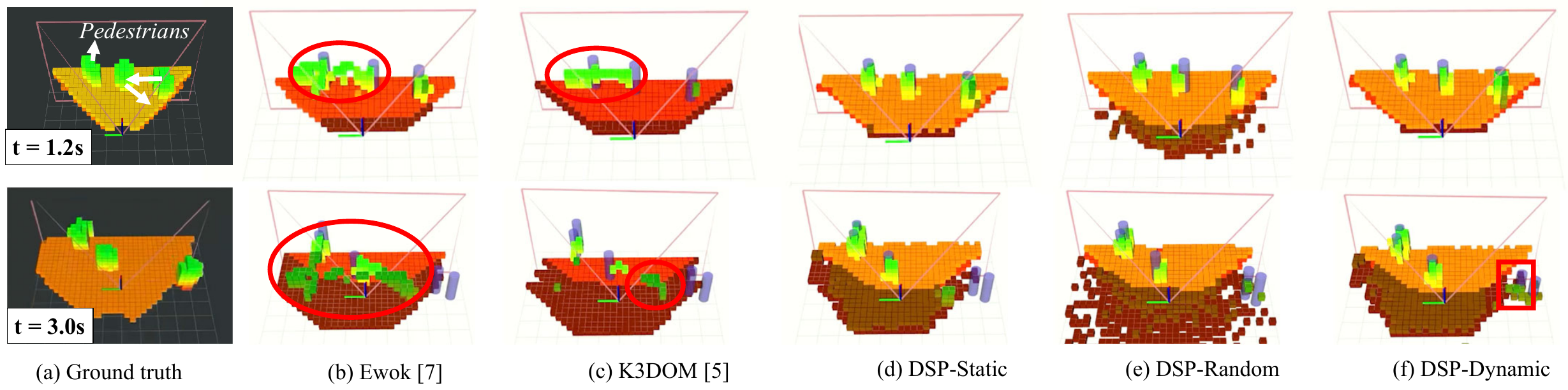,width=7.1in}}
  \caption{Snapshots of different maps in the pedestrian square world when the mapping resolution is 0.3 m. The red ellipses show the trail noise caused by moving pedestrians in Ewok \cite{Ringbuffer} and K3DOM \cite{DynamicMapICRA2021}. From $t=1.2$s to $t=2.8$s, the movement of three pedestrians causes more trail noise. The red rectangle in Column (f) shows the predicted occupancy of a pedestrian out of FOV.} 
  \label{Fig: mapping_performance}
\end{figure*}

K3DOM runs on NVIDIA RTX 2060 GPU, and the rest maps run on AMD Ryzen 4800HS CPU in the tests. The map size $(l_x, l_y, l_z)$ is (10 m, 10 m, 6 m). The rest parameters in K3DOM and Ewok remain the same as the original settings in the released code. No voxel filter is used for point cloud pre-processing in K3DOM and Ewok to reach their best performances. In DSP map and its variants, the initial weight of the particle is $0.0001$. Three different voxel sizes, from 0.1 m to 0.3 m, were tested in the simulation worlds shown in Fig. \ref{Fig: testing scenarios} (b) to (d). Using different occupancy probability thresholds, which determine the binary status, i.e., occupied or free, we draw precision-recall curves in Fig \ref{Fig: accuracy_recall_curve_large} (b). 
Snapshots of different maps can be found in Fig. \ref{Fig: mapping_performance_high_resolution} and Fig. \ref{Fig: mapping_performance}.

In Fig \ref{Fig: accuracy_recall_curve_large} (b), a larger area under the curve (AUC) suggests a better overall performance in classifying the occupancy status with different thresholds. The specific AUC values are presented in Table \ref{Table: AUC}. 
The DSP-Dynamic map has the largest AUC in the two worlds with pedestrians and a comparable AUC with Ewok and DSP static map in the static forest world. 
When the voxel size is 0.1 m and 0.2 m, the recall of Ewok and K3DOM is relatively low because of the gaps and inconsistencies in high-resolution grid maps.
Red dashed boxes in Fig. \ref{Fig: mapping_performance_high_resolution} illustrate the gaps and inconsistencies.
In Fig. \ref{Fig: mapping_performance}, the areas in the red ellipses show that Ewok has noticeable trail noise, which can lower the precision, when the voxel size is 0.3 m. The dynamic occupancy map K3DOM has less trail noise.
In comparison, our DSP-Dynamic map doesn't suffer from gaps, inconsistencies, or trail noise.

DSP-Random, which doesn't have initial velocity estimation and uses only the CV model, has obvious noise in the area out of FOV, especially when representing static obstacles. Column (e) in Fig. \ref{Fig: mapping_performance_high_resolution} shows the noise. Consequently, the AUC of DSP-Random is the smallest in the forest world. DSP-Static adopts only the static motion model and achieves the best AUC in the forest world when the voxel size is 0.2 m and 0.3 m. However, it cannot predict the future occupancy status of dynamic obstacles, and the AUCs in the worlds with pedestrians are smaller than DSP-Dynamic. The red rectangles in Column (f) in Fig. \ref{Fig: mapping_performance_high_resolution} and Fig. \ref{Fig: mapping_performance} show the predicted occupancy status of a pedestrian out of the FOV in the DSP-Dynamic map.

Table \ref{Table: F1-Score} shows the best F1-Score, i.e., the highest classification performance that each map reaches with different probability thresholds in Fig \ref{Fig: accuracy_recall_curve_large} (b).
When the testing scenario is the forest world, and the voxel size is 0.3 m, the DSP-Dynamic map's score is slightly lower than the DSP-Static map's. In all other situations, DSP-Dynamic has the highest best F1-Score. 
Note in the pedestrian square world, where only dynamic obstacles exist, the dynamic map K3DOM has a lower best F1-Score than the static map Ewok when the voxel size is 0.3 m. The reason is that although Ewok has a low precision due to its heavy trail noise, its recall rate is higher than K3DOM's. 
However, from Table \ref{Table: AUC}, it can be seen that the AUC, which evaluates the overall classification performance when using different occupancy probability thresholds, of K3DOM is still higher than that of Ewok.

The average F1-score and AUC of our DSP-Dynamic map in different worlds with different resolutions are 0.46 and 0.47, respectively. In comparison, the average F1-score and AUC of the existing particle-based dynamic occupancy map K3DOM are 0.33 and 0.37, respectively. Our map increases the F1-score by 39.4\% and AUC by 27.0\%. If only the two worlds that contain dynamic obstacles are considered, the average F1-score and AUC increase from 0.24 to 0.39 (62.5\% increase) and 0.27 to 0.40 (48.1\% increase), respectively.

\begin{table}[h]\footnotesize 
\center
\caption{AUC Comparison}
\label{Table: AUC}
\resizebox{\linewidth}{!}{
\begin{tabular}{l|lll|lll|lll}
\hline
World                                            & \multicolumn{3}{c|}{Pedestrian square}                                                           & \multicolumn{3}{c|}{Forest}                                                                      & \multicolumn{3}{c}{Pedestrian street}                                                            \\ \hline
Voxel size (m)                                    & 0.1                            & 0.2                            & 0.3                            & 0.1                            & 0.2                            & 0.3                            & 0.1                            & 0.2                            & 0.3                            \\ \hline
Ewok \cite{Ringbuffer}          & 0.10                           & 0.22                           & 0.27                           & 0.59                           & 0.59                           & 0.66                           & 0.28                           & 0.34                           & 0.46                           \\
K3DOM \cite{DynamicMapICRA2021} & 0.15                           & 0.23                           & 0.28                           & 0.55                           & 0.57                           & 0.61                           & 0.27                           & 0.33                           & 0.37                           \\
DSP-Random                                       & 0.24                           & 0.24                           & 0.36                           & 0.32                           & 0.35                           & 0.50                           & 0.27                           & 0.31                           & 0.44                           \\
DSP-Static                                       & 0.24                           & 0.29                           & 0.32                           & 0.56                           & \textbf{0.61} & \textbf{0.69} & 0.34                           & 0.42                           & 0.49                           \\
DSP-Dynamic                                      & \textbf{0.29} & \textbf{0.36} & \textbf{0.43} & \textbf{0.60} & 0.59                           & 0.65                           & \textbf{0.36} & \textbf{0.43} & \textbf{0.50} \\ \hline
\end{tabular}}
\end{table}

\begin{table}[h]\footnotesize 
  \center
  \caption{Best F1-Score Comparison}
  \label{Table: F1-Score}
  \resizebox{\linewidth}{!}{
  \begin{tabular}{l|lll|lll|lll}
  \hline
  World          & \multicolumn{3}{c|}{Pedestrian square}           & \multicolumn{3}{c|}{Forest}                      & \multicolumn{3}{c}{Pedestrian street}                       \\ \hline
  Voxel size (m)      & 0.1            & 0.2            & 0.3            & 0.1            & 0.2            & 0.3            & 0.1            & 0.2            & 0.3            \\ \hline
  Ewok \cite{Ringbuffer}        & 0.07          & 0.19          & 0.24          & 0.42          & 0.50          & 0.52          & 0.24          & 0.32          & 0.40          \\
  K3DOM \cite{DynamicMapICRA2021}      & 0.14          & 0.19          & 0.22          & 0.53          & 0.53          & 0.44          & 0.27          & 0.32          & 0.32          \\
  DSP-Random  & 0.24          & 0.24          & 0.35          & 0.29          & 0.29          & 0.44          & 0.25          & 0.28          & 0.39          \\
  DSP-Static  & 0.17          & 0.24          & 0.29          & 0.48          & \textbf{0.59}          & \textbf{0.66} & 0.30          & \textbf{0.41}          & 0.49          \\
  DSP-Dynamic & \textbf{0.26} & \textbf{0.36} & \textbf{0.43} & \textbf{0.58} & \textbf{0.59} & 0.65          & \textbf{0.36} & \textbf{0.41} & \textbf{0.50} \\ \hline
  \end{tabular}}
  \end{table}

\subsection{Robotics Platform Efficiency Tests}
\begin{figure}[h]
  \centerline{\psfig{figure=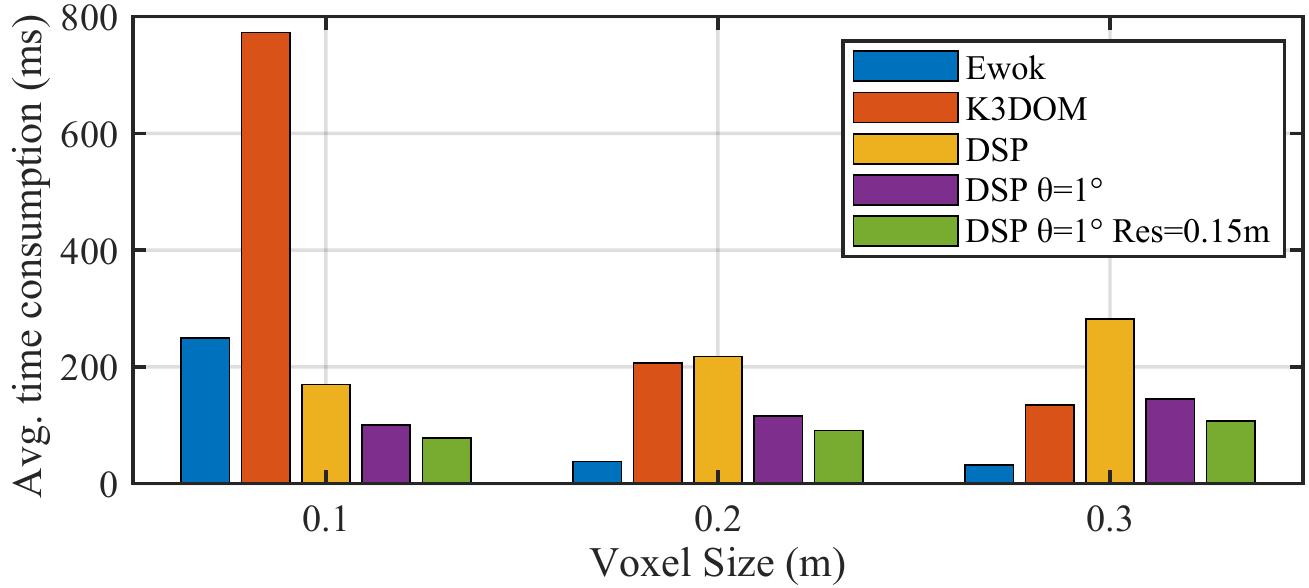, width=2.8in}}
  \caption{Time consumption of different maps on Jeston Xavier NX.}
  \label{Fig: nx time}
\end{figure}

This section first compares the efficiency of Ewok \cite{Ringbuffer}, K3DOM \cite{DynamicMapICRA2021}, and our DSP map (with particle initial velocity estimation and mixture motion model) on NVIDIA Jetson Xavier NX, which is a small computing board widely used on robotics platforms. Xavier NX has a 384-core NVIDIA Volta GPU with 48 Tensor Cores and a 6-core NVIDIA Carmel ARM v8.2 CPU.
K3DOM \cite{DynamicMapICRA2021} runs on the GPU, and the rest maps run on the CPU. The average time consumption of each map with different voxel sizes is shown in Fig. \ref{Fig: nx time}. The map size in the test is (10 m, 10 m, 6 m). 

When the voxel size is 0.1 m, our DSP map is the fastest. The existing particle-based dynamic occupancy map K3DOM is 4.5 times slower than the DSP map. The static map Ewok runs fastest when the voxel size is 0.2 m or 0.3 m. K3DOM is the second fastest, and our DSP map is the slowest. However, with the results in Fig. \ref{Fig: accuracy_recall_curve_large} (a), we can further raise the computational efficiency of our map by sacrificing a little performance on the F1-Score. 
Fig. \ref{Fig: accuracy_recall_curve_large} (a) indicates that decreasing the pyramid subspace angle $\theta$ from $3^\circ$ to $1^\circ$ can reduce the computation time while the F1-Score drops merely 1\%. 
In addition, increasing the filter voxel size $Res$ from 0.1 m to 0.15 m can also reduce the computation time, and the F1-Score decreases 12\% accordingly. 
If $\theta = 1^\circ$ is used in the tests on Xavier NX, the DSP map's computation time is only 0.56 times that of K3DOM's when the voxel size is 0.2 m and is close to K3DOM's when the voxel size is 0.3 m.
If $Res = 0.15 m$ is further adopted, the DSP map's computation time is shorter than the K3DOM's for all tested voxel sizes.

We also tested the computation efficiency of our DSP map on two other onboard computers for robotics platforms: an Intel NUC with a Core i7-10710u CPU and an Up core board with an Intel Atom x5-z8350 CPU. When $\theta = 1^\circ$ and $Res = 0.15$ m are adopted, and the voxel size is $0.2$ m, the average time consumption on the two boards is about 133 ms and 254 ms, respectively. For robotics obstacle avoidance tasks without fast movement, a smaller map can be used to reduce time consumption. For example, when the map size is reduced to (8 m, 8 m, 3 m), the average time consumption on the Up core board is below 150 ms.

In Appendix \ref{App: Test with Lidar Input}, we present a test with omnidirectional and multi-channel Lidar point cloud data. The time consumption is about two times when using point cloud data from the Realsense camera, which has a limited FOV. Improvements in computational efficiency will be conducted further to realize real-time mapping with multi-channel Lidars.


\subsection{Applications}
\begin{figure*}[h]
\centerline{\psfig{figure=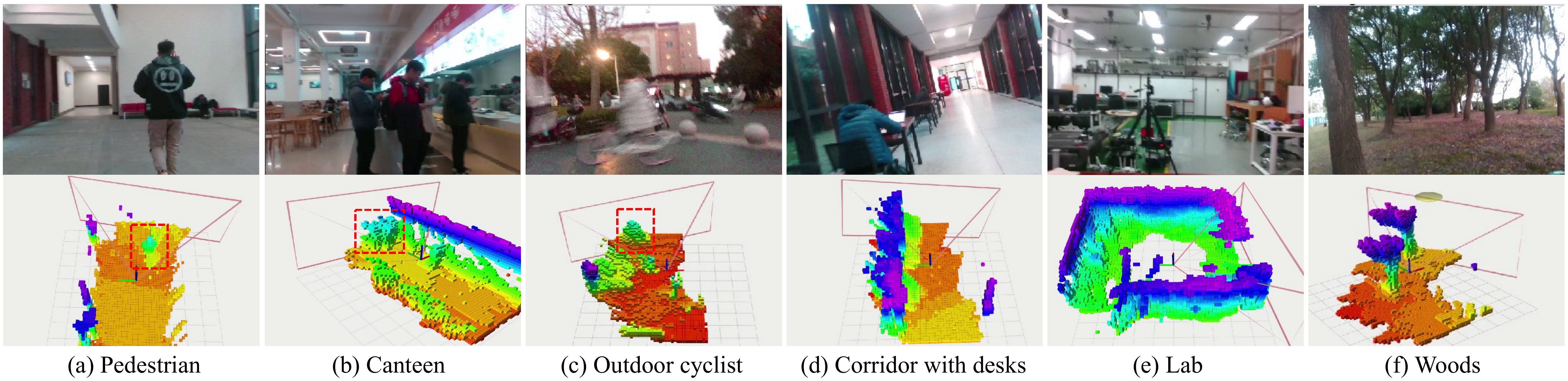,width=7.1in}}
\caption{Snapshots of building the DSP-Dynamic map in different scenarios. The first row shows the RGB image in current FOV. The second row presents the voxelized map view with a resolution of 0.15 m. The pink outlines show the FOV. Red dashed boxes indicate the dynamic obstacles in current FOV.}
\label{Fig: real world tests}
\end{figure*}

\begin{figure}[h]
  \centerline{\psfig{figure=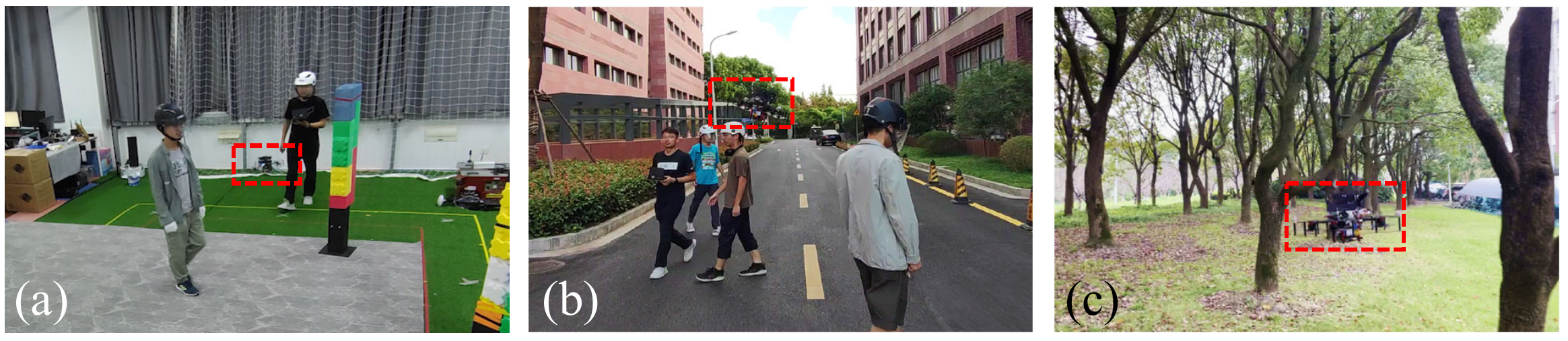, width=3.4in}}
  \caption{The testing scenarios for obstacle avoidance. (a) and (b) are dynamic environments. (c) is a static environment. Red rectangles outline the quadrotor.}
  \label{Fig: mav_test}
\end{figure}

Fig. \ref{Fig: real world tests} presents several snapshots of building the DSP-Dynamic map in different scenarios. The localization was realized by a Realsense T265 tracking camera, and the point cloud was from a Realsense d435 camera.
To further demonstrate the effectiveness and efficiency of our map in robotic systems. We deployed the DSP-Dynamic map on a mini quadrotor with a weight of only 320 grams and utilized a sampling-based motion planning method \cite{OursDynamicPlanning} to realize obstacle avoidance in environments with static and dynamic obstacles. The method samples motion primitives and evaluates the collision risk of each motion primitive with the current and predicted particles in the DSP-Dynamic map. Details can be found in \cite{OursDynamicPlanning}.
The point cloud was collected from a Realsense d435 camera, and everything, including mapping and motion planning, was performed on the CPU of a low-cost Up core computing board. Fig. \ref{Fig: mav_test} shows the testing scenarios. The testing demos can be found at \url{https://youtu.be/seF_Oy4YbXo}. 

\section{Conclusion} \label{Section: conclusion}
This paper presents a novel dual-structure particle-based 3D local map, named DSP (dynamic) map, that allows continuous occupancy mapping of dynamic environments. Voxel subspaces and pyramid-like subspaces are adopted to achieve efficient updates in continuous space. The initial velocity estimation and a mixture model are considered to reduce noise. Experiments show that the DSP map can increase the dynamic obstacle velocity estimation performance by over 30\% on MBD, compared to other tested point-cloud-based methods.
In occupancy status estimation tests, the DSP map increases the F1-Score of the state-of-the-art particle-based occupancy map from 0.33 to 0.46 (39.4\% increase) and the AUC from 0.37 to 0.47 (27.0\% increase) on average.
Furthermore, efficiency tests and a real-world application demo demonstrated the broad prospect of this map in obstacle avoidance tasks of small-scale robotic systems.
Future works will consider two main points. The first is to introduce semantic information to this map to better identify and model different obstacles and further predict their future states with multiple hypotheses. The second is to connect this dynamic local map to a global static map to achieve global mapping in dynamic environments.




{\appendices

\section{Lower Bound Distance Calculation} \label{App: Lower Bound Calculation}
This appendix calculates the lower bound distance from a point object to a measurement point whose azimuth angle and zenith angle differences with the point object are no less than $\theta^\prime$. Fig. \ref{Fig: Lower Bound Calculation} shows two limiting cases where the zenith and azimuth angle difference are $\theta^\prime$, respectively. In subplot (a), when $\boldsymbol{P}_{\boldsymbol{z}_k}$ is in the same vertical plane with $\boldsymbol{P}_{\boldsymbol{x}_k}$ and $\boldsymbol{P}_{\boldsymbol{x}_k}\boldsymbol{P}_{\boldsymbol{z}_k}  \perp  \boldsymbol{P}_{\boldsymbol{z}_k}\boldsymbol{O}$, the minimum distance between $\boldsymbol{P}_{\boldsymbol{x}_k}$ and $\boldsymbol{P}_{\boldsymbol{z}_k}$ exists and is $|\boldsymbol{P}_{\boldsymbol{x}_k}\boldsymbol{P}_{\boldsymbol{z}_k}| = r_k sin \theta^\prime$. 
In Subplot (b), $|\boldsymbol{P}_{\boldsymbol{x}_k}\boldsymbol{P}_{\boldsymbol{z}_k}| \geq |\boldsymbol{P}_{\boldsymbol{x}_k}^\prime \boldsymbol{P}_{\boldsymbol{z}_k}^\prime|$, where $|\boldsymbol{P}_{\boldsymbol{x}_k}^\prime \boldsymbol{P}_{\boldsymbol{z}_k}^\prime|$ is the distance between the projection points of $\boldsymbol{P}_{\boldsymbol{x}_k}$ and $\boldsymbol{P}_{\boldsymbol{z}_k}$. When $\boldsymbol{P}_{\boldsymbol{x}_k}^\prime \boldsymbol{P}_{\boldsymbol{z}_k}^\prime  \perp  \boldsymbol{P}_{\boldsymbol{z}_k}^\prime \boldsymbol{O}$, $|\boldsymbol{P}_{\boldsymbol{x}_k}^\prime \boldsymbol{P}_{\boldsymbol{z}_k}^\prime|$ has the minimum value $r_k sin \alpha_k sin \theta^\prime$. Since $r_k sin \alpha_k sin \theta^\prime \leq r_k sin \theta^\prime$, the lower bound of $|\boldsymbol{P}_{\boldsymbol{x}_k}\boldsymbol{P}_{\boldsymbol{z}_k}|$ is $r_k sin \alpha_k sin \theta^\prime$.

\begin{figure}[h]
  \centerline{\psfig{figure=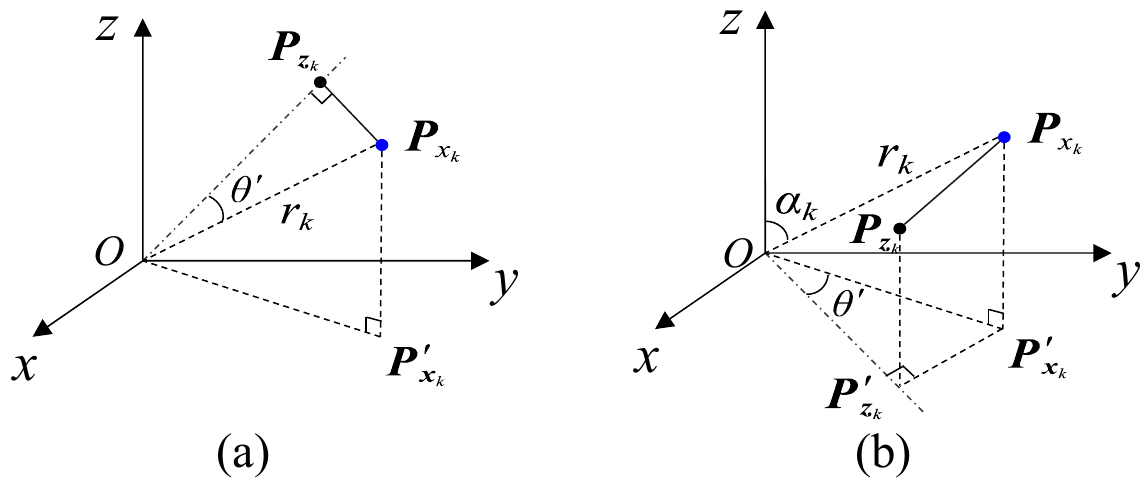, width=2.5in}}
  \caption{Calculation of the lower bound distance from a point object position $\boldsymbol{P}_{\boldsymbol{x}_k}$ to a measurement point the position $\boldsymbol{P}_{\boldsymbol{z}_k}$. The azimuth angle and zenith angle difference between $\boldsymbol{P}_{\boldsymbol{z}_k}$ and $\boldsymbol{P}_{\boldsymbol{x}_k}$ is no less than $\theta^\prime$. In subplot (a),the zenith angle difference is $\theta^\prime$. In subplot (a), the azimuth angle difference is $\theta^\prime$. $\boldsymbol{O}$ is the origin point. $\alpha_k$ is the zenith angle of $\boldsymbol{P}_{\boldsymbol{x}_k}$. $r_k$ is the distance from $\boldsymbol{O}$ to $\boldsymbol{P}_{\boldsymbol{x}_k}$. $\boldsymbol{P}_{\boldsymbol{x}_k}^\prime$ and $\boldsymbol{P}_{\boldsymbol{z}_k}^\prime$ are the projection point of $\boldsymbol{P}_{\boldsymbol{x}_k}$ and $\boldsymbol{P}_{\boldsymbol{z}_k}$ in the $x-y$ plane, respectively.}
  \label{Fig: Lower Bound Calculation}
\end{figure}

\section{Test with Lidar Input} \label{App: Test with Lidar Input}
\begin{figure}[h]
  \centerline{\psfig{figure=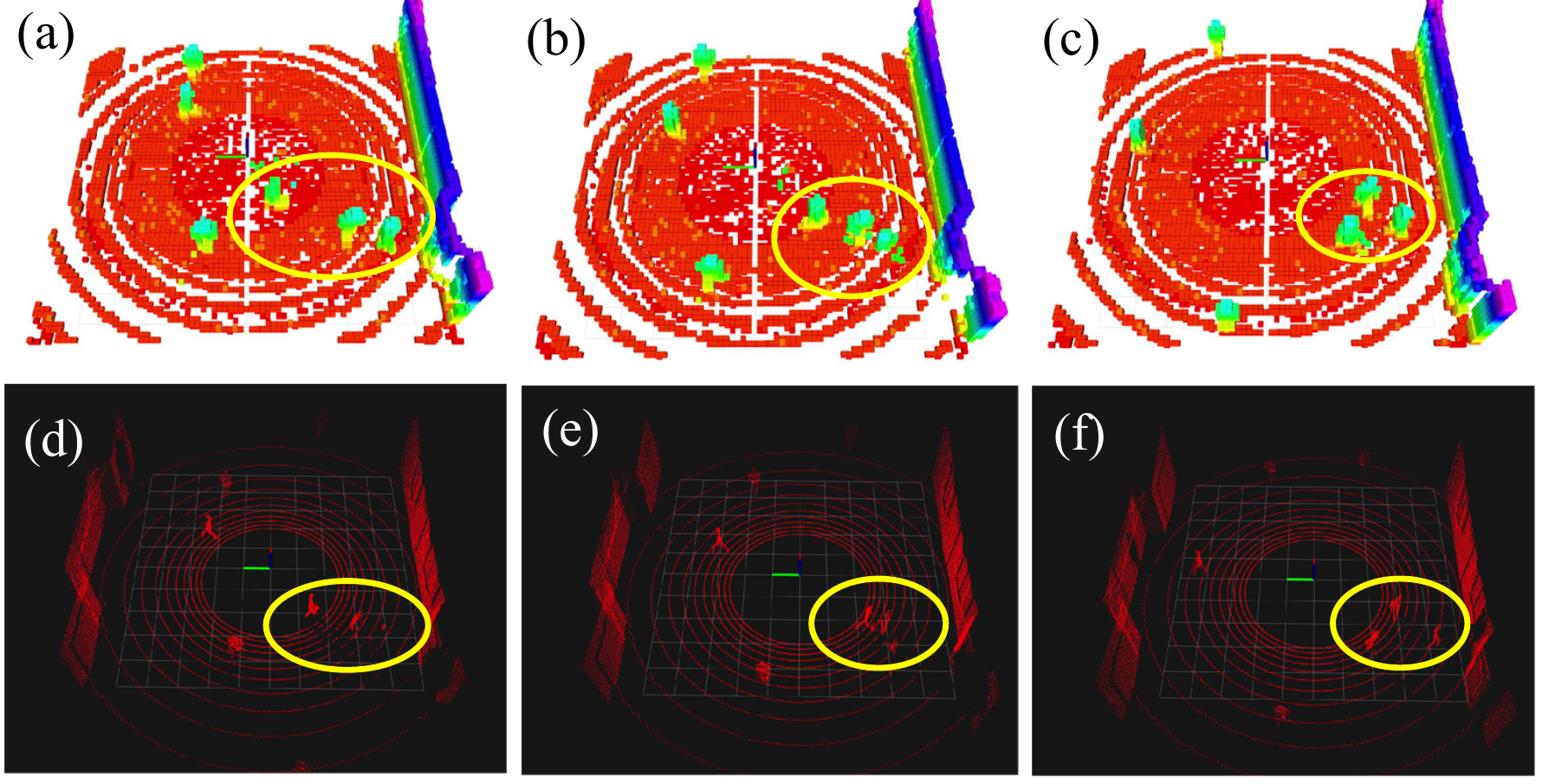, width=3.0in}}
  \caption{Snapshots of mapping with the point cloud from Lidar. The top row shows the DSP map, and the bottom row shows the point cloud data. The yellow ellipses outline three pedestrians. In (d) and (e), the two pedestrians on the right are occluded or partially occluded but are clearly shown in the map. In (f), the pedestrians are detected by the Lidar again.}
  \label{Fig: lidarMap}
\end{figure}

This appendix presents a qualitative test result with point cloud data from a simulated Velodyne HDL-32E Lidar. This Lidar is an omnidirectional and 32-channel Lidar with a horizontal resolution of $0.16^\circ$ and a vertical resolution of $1.33^\circ$. The FOV is $360^\circ \times 40^\circ$. The point cloud data is collected in the pedestrian street world shown in Fig. \ref{Fig: testing scenarios} (d). The mapping parameters are $\theta=3^\circ$, $Res=0.15m$, and $L_{max} = 1.6 \times 10^6$. The map size is (10 m, 10 m, 6 m). Fig.\ref{Fig: lidarMap} shows several snapshots of the mapping result. 

When the voxel size for discrete occupancy status estimation is 0.1 m, 0.2 m, and 0.3 m, the time consumption for mapping is 194.9 ms, 259.4 ms, and 470.4 ms, respectively. In Section \ref{Section: Update}, we have discussed that the complexity in the update procedure is $O(\frac{\theta^2}{\theta_h\theta_v} L_k M_k)$. The omnidirectional character of the used Lidar increases $\theta_h \theta_v$ but also increases the particle number $M_k$ in $\mathbb{M}^f$. The result is that the time consumption is about two times the time consumption of using a depth camera with a smaller FOV ($90^\circ \times 60^\circ$). This time consumption is not small enough for real-time usage. Further improvements in computational efficiency will be conducted to realize real-time mapping with this kind of Lidar.

\section{Unknown Area Representation}
Representing the unknown area is very useful in exploration tasks. For static grid maps, the grids are initialized with a tag ``unknown'', and the tag is removed when a ray generated from point cloud measurement passes through or hits the grid. In the DSP map, the unknown area can be represented by the update time of the particles. Adding time stamps on the common particles doesn't work because the particles are born only in the area with obstacles, and thus the unknown area and the free area cannot be distinguished. Therefore, when a new area appears in the map, a small number of static particles, named time particles, which have a zero weight and a time stamp, can be uniformly added to the map. When the measurement point cloud comes, the time particles only update their timestamp to the current time. Then the unknown property of each area can be evaluated by checking the time stamp. 


\section{Data Structure Details} \label{Appendix: Algorithms for Data Structure}
Algorithm \ref{Algorithm: Voxel Particle} shows the data structure to store particles in voxel subspaces and basic operations used in the mapping algorithms. $ptc\_voxel\_array$ is a fixed-size array that stores particles in voxels. The maximum particle number that can be stored in a voxel subspace is $\frac{\eta_1 L_{max}}{N_v}$, where $\eta_1 > 1$ is an empirical factor used to allocate large storage space.
If no storage space is left in the array, the particle to be added is omitted. 
Note $\frac{\eta_1 L_{max}}{N_v}$ is larger than $L_{max}^{\mathbb V}$ used in the resampling step because, in the prediction and particle birth steps, more than $L_{max}^{\mathbb V}$ particles may enter one voxel subspace and should all be stored. In the resampling step, the number of particles is reduced to $L_{max}^{\mathbb V}$ with Eq. (\ref{Eq: resampling_nums}).

Algorithm \ref{Algorithm: Pyramid Particle} illustrates how the indexes of particles are stored in the pyramid subspaces. $vr\_ptc\_pyd\_array$ is the array to store particle indexes in pyramid subspaces. $\eta_2 > 1$ is another empirical factor that works similar to $\eta_1$. In practice, $\eta_1 = \eta_2 = 3$ is adopted. 
Note that the pyramid subspaces are divided dynamically with the sensor's orientation, and thus, all the particle indexes in $vr\_ptc\_pyd\_array$ must be updated in real-time. Therefore, $vr\_ptc\_pyd\_array$ is emptied after each mapping process and recalculated in the prediction step (Algorithm \ref{Algorithm: Prediction}).

\begin{algorithm*}[h]
  \caption{\small Particle Storage and Operation in Voxel Subspaces}
  \footnotesize
  \SetKwProg{Fn}{Function}{}{end}
  \label{Algorithm: Voxel Particle}
  \textbf{Struct} Particle $Particle = \left\{flag, weight, v_x, v_y, v_z, p_x, p_y, p_z \right\}$  \algorithmiccomment{Struct of the states of a particle. $flag$ is initialized with 0}\;
  $L_{s}^{\mathbb V} \gets \eta_1 \cdot L_{max} / N_v$ 
  \algorithmiccomment{Max particle number stored in a voxel. $\eta_1 > 1$ is an empirical factor. $N_v=\frac{l_x l_y l_z}{l^3}$. $L_{max} / N_v = L_{max}^{\mathbb V}$}\; 
  $ptc\_voxel\_array[N_v][L_{s}^{\mathbb V}]\{Particle\}$ \algorithmiccomment{This array stores the particles in voxel subspaces. $N_v$ is the number of voxel subspaces}\;
  \Fn{\rm addParticleToVoxel($Particle$)}{
    $idx\_voxel \gets $getIdxVoxel$(Particle.p_x, Particle.p_y, Particle.p_z)$ \algorithmiccomment{Calculate the voxel index of a particle with the particle's position}\; 
    $i \gets$ getVacancyIdx$(ptc\_voxel\_array[idx\_voxel])$ \algorithmiccomment{Traverse $ptc\_voxel\_array[idx\_voxel]$ and check the $flag$ until a vacancy with $flag=0$ is found}\;
    $Particle.flag \gets 1$, $ptc\_voxel\_array[idx\_voxel][i] \gets Particle$ \algorithmiccomment{Add $Particle$ to the vacancy in $ptc\_voxel\_array$}\; 
    \textbf{return} $idx\_voxel, \ i$\;
  }
  \Fn{\rm deleteParticleInVoxel($idx\_voxel, idx\_ptc$)}{
    $ptc\_voxel\_array[idx\_voxel][idx\_ptc].flag \gets 0$ \algorithmiccomment{Delete a particle by setting its flag to zero}\; 
  }
  \Fn{\rm moveParticleToNewVoxel($idx\_voxel, idx\_ptc$)}{
    deleteParticleInVoxel($idx\_voxel, idx\_ptc$) \algorithmiccomment{Delete the particle in the original voxel}\; 
    \textbf{return} addParticleToVoxel($ptc\_voxel\_array[idx\_voxel][idx\_ptc]$) \algorithmiccomment{Add the particle to a new voxel and return indexes}\; 
  }
  \end{algorithm*}

  \begin{algorithm*}[h]
    \footnotesize
    \caption{\small Particle Index Storage and Operation in Pyramid Subspaces}
    \SetKwProg{Fn}{Function}{}{end}
    \SetKwProg{Struct}{Struct}{}{end}
    \label{Algorithm: Pyramid Particle}
    \textbf{Struct} Virtual\_Particle $Vr\_Particle = \left\{flag, idx\_voxel, idx\_ptc \right\}$  \algorithmiccomment{Struct of a virtual particle, which contains indexes mapping to the particle in $ptc\_voxel\_array[idx\_voxel][idx\_ptc]$. $flag$ is initialized with 0}\;
    $L_{s}^{A} \gets \eta_2 \cdot \frac{L_{max} \cdot \theta^2}{360 \times 180}$ \algorithmiccomment{Max particle number in a pyramid. $\eta_2 > 1$ is an empirical factor}\;
    $vr\_ptc\_pyd\_array[N_f][L_{s}^{A}]\left\{ Vr\_Particle \right\}$ \algorithmiccomment{This array stores the Virtual\_Particles in each pyramid}\;
    $pyd\_neighbors\_array[N_f][n^2-1]$ \algorithmiccomment{This array stores the neighbor pyramids' indexes in the activation range. $n$ is defined in Section \ref{Section: Update}}\;
    
    \Fn{\rm getIdxPyd($p_x, p_y, p_z$) \algorithmiccomment{This function calculates the pyramid index of a particle or a point.}}{ 
      $\{p_{x,s}, p_{y,s}, p_{z,s}\} \gets \text{worldToSensorFrame}(p_x, p_y, p_z)$   \algorithmiccomment{Calculate the position of a particle in sensor frame}\;
      \If{\rm pointInFov($p_{x,s}, p_{y,s}, p_{z,s}$)}{
        $idx\_pyd \gets$ getIdxPydSensorFrame$(p_{x,s}, p_{y,s}, p_{z,s})$ \algorithmiccomment{Calculate the pyramid index of a point using sensor frame position}\;
        \textbf{return} $idx\_pyd$\;
      } 
      \textbf{else} \textbf{return} -1\;
    }

    \Fn{\rm addParticleIdxToPyd($idx\_pyd,idx\_voxel,idx\_ptc$) \algorithmiccomment{This function adds the index of a particle (a virtual particle) to $vr\_ptc\_pyd\_array$.}}{ 
      $j \gets$ getVacancyIdx$(vr\_ptc\_pyd\_array[idx\_pyd])$ \algorithmiccomment{Traverse $vr\_ptc\_pyd\_array[idx\_pyd]$ until a vacancy ($flag=0$) is found}\;
      $vr\_ptc\_pyd\_array[idx\_pyd][j] \gets \left\{1,idx\_voxel,idx\_ptc \right\}$ \algorithmiccomment{Add the particle index to the vacancy}\;
    }
\end{algorithm*}

\section{Algorithms for Mapping} \label{Appendix: Algorithms for Mapping}
The algorithms to realize DSP map building are illustrated in Algorithm \ref{Algorithm: Preprocess} to \ref{Algorithm: New-born}. In practice, we use C++ for programming. 
Algorithms \ref{Algorithm: Prediction} to \ref{Algorithm: New-born} correspond to the prediction, update, particle birth and resampling steps in Section \ref{Section: Map Building}, respectively.
These algorithms show one way to implement the mapping methods.
The computational complexities of Algorithm \ref{Algorithm: Prediction}, \ref{Algorithm: Update}, \ref{Algorithm: Resample and Occupancy Status Estimation} and \ref{Algorithm: New-born} are $O(L_{max})$, $O(M_k L_{max} \theta^2)$, $O(L_{max})$, and $O(M_k)$, respectively. The overall computational complexity of the map is then $O(M_k L_{max} \theta^2)$. Note the lookup operation in Function $getVacancyIdx$ is disregarded in the computational complexity analysis to simplify the expression. The lookup operation is confined to a small subspace and costs little computing resource. 


\begin{algorithm*}
  \footnotesize
  \caption{\small Point Cloud Preprocess}
  \SetKwProg{Fn}{Function}{}{end}
  \SetKwProg{Struct}{Struct}{}{end}
  \label{Algorithm: Preprocess}
  
  $ptcl\_array[M_k]\{p_x,p_y,p_z\}$ \algorithmiccomment{The input array that stores the point cloud filtered by a voxel filter with resolution $Res$}\;
  $M^{A}_s \gets \frac{\theta^2}{\theta_{snsr}^2}$  \algorithmiccomment{Calculates the maximum measurement point stored in a pyramid. $\theta_{snsr}$ is the angle resolution of the sensor}\; 
  $ptcl\_pyd\_array[N_f][M^{A}_s]\{p_x,p_y,p_z\}$ \algorithmiccomment{This array stores the point cloud in each pyramid in the map frame}\; 
  $pyd\_pt\_num\_array[N_f]$, $pyd\_length\_array[N_f]$
  \algorithmiccomment{The arrary of points number and the visible length of each pyramid}\;
  \For{\rm $i=1:M_k$}{
    $pt \gets ptcl\_array[i]$,
    $idx\_pyd \gets \text{getIdxPyd}(pt.p_x, pt.p_y, pt.p_z)$\algorithmiccomment{Get the pyramid subspace index of a point}\; 

    rotateAndStore$(ptcl\_pyd\_array, pt, sensor\_ort)$ \algorithmiccomment{Rotate $pt$ to the map frame with sensors' orientation and store $pt$ in $ptcl\_pyd\_array$}\;

    $pyd\_pt\_num\_array[idx\_pyd] ++$ \algorithmiccomment{Update the number of measurement points in a pyramid}\; 
    $eu\_dist \gets$ squareEuclideanDist$(pt)$ \algorithmiccomment{Calculate square Euclidean distance from $pt$ to the map center}\;
    \uIf{$eu\_dist > pyd\_length\_array[idx\_pyd]$}{
      $pyd\_length\_array[idx\_pyd] \gets eu\_dist$ \algorithmiccomment{Update the visible length of a pyramid in the FOV}\;
    }
  }
  \end{algorithm*}

  \begin{algorithm*}
    \footnotesize
    \caption{\small Particle Initial Velocity Estimation}
    \SetKwProg{Fn}{Function}{}{end}
    \SetKwProg{Struct}{Struct}{}{end}
    \label{Algorithm: Initial Velocity Estimation}
    $ptcl\_vel\_array[N_f][M^{A}_s]\{p_x,p_y,p_z\}$ \algorithmiccomment{This array stores the estimated velocities of points in the point cloud}\;
    $ptcl\_vel\_array = $ calPointVelocity($ptcl\_array$)   
    \algorithmiccomment{Calculate $ptcl\_vel\_array$ with procedures in Fig. \ref{Fig: velocity estimation} (b)}\; 
  \end{algorithm*}

  \begin{algorithm*}
  \footnotesize
  \caption{\small Prediction Step}
  \SetKwProg{Fn}{Function}{}{end}
  \SetKwProg{Struct}{Struct}{}{end}
  \label{Algorithm: Prediction}
  Empty $vr\_ptc\_pyd\_array[N_f][L_{s}^{A}]\left\{ Vr\_Particle \right\}$ \algorithmiccomment{Empty $vr\_ptc\_pyd\_array$ and recalculate later to realize dynamical division of the pyramid subspaces}\;
  $pyd\_vr\_ptc\_num\_array[N_f]$ \algorithmiccomment{The array to store the number of particle indexes in a pyramid}\;
  \For{\rm $i$ in Range($N_v$)}{
    \For{\rm $j$ in Range($L_{s}^{\mathbb V}$)}{
      \uIf{$ptc\_voxel\_array[i][j].flag$ is 0}{
        \textbf{continue} \algorithmiccomment{Ignore the array element that doesn't contain a particle currently}\;
      }
      $particle \gets $predictParticleState($ptc\_voxel\_array[i][j]$) \algorithmiccomment{Predict the particle's state with the mixture model in Eq. (\ref{Eq: single_object_motion}) and (\ref{Eq: mixture model sample}).}
  
      $idx\_voxel, idx\_ptc \gets$moveParticleToNewVoxel($i,j$) \algorithmiccomment{Move the particle with the predicted state}\;
  
      $idx\_pyd \gets$getIdxPyd($particle.x, particle.y, particle.z$)
      \algorithmiccomment{Get the pyramid index of the particle after prediction}\;
      \uIf{$idx\_pyd \geq 0$}{
        addParticleIdxToPyd($idx\_pyd,idx\_voxel, idx\_ptc$)\algorithmiccomment{Add the particle index to $vr\_ptc\_pyd\_array$ for update}\;
        $pyd\_vr\_ptc\_num\_array[idx\_pyd] ++$ \algorithmiccomment{Update the number of particle indexes in the pyramid}\;
      }
    }
  }
  
  \end{algorithm*}

  \begin{algorithm*}[t]
  \footnotesize
  \caption{\small Update Step}
  \SetKwProg{Fn}{Function}{}{end}
  \SetKwProg{Struct}{Struct}{}{end}
  \label{Algorithm: Update}
  
  $\text{C}_k^\prime\_array[N_f][M^{A}_s]$ \algorithmiccomment{This array stores variable $\text{C}_k^\prime(\boldsymbol{z}_k)$ for each $\boldsymbol{z}_k$ and is defined in Eq. (\ref{Eq: particles_posterior_weights3_approx_newborn})}\;
  
  \For{$i=1:N_f$}{
    \For{$j=1:pyd\_pt\_num\_array[i]$}{
      $pt \gets ptcl\_pyd\_array[i][j]$ \algorithmiccomment{Get the point in the point clould one by one}\; 
      $\text{C}_k^\prime\_array[i][j] \gets $calculateCk($pt$,$ vr\_ptc\_pyd\_array$, $pyd\_neighbors\_array$) \algorithmiccomment{Calculate $\text{C}_k^\prime$ with Eq. (\ref{Eq: particles_posterior_weights3_approx_newborn})}\;
    }
  }
  \For{$i=1:N_f$}{
    \For{$j=1: pyd\_vr\_ptc\_num\_array[i]$}{
      $ptc \gets ptc\_voxel\_array\left[vr\_ptc\_pyd\_array[i][j].id\_voxel\right]\left[vr\_ptc\_pyd\_array[i][j].idx\_ptc\right]$ \algorithmiccomment{Get the particle to update}\;
      $ptc\_dist \gets squareEuclideanDist(ptc)$ \algorithmiccomment{Calculate square Euclidean distance from $ptc$ to the map center}\;
      \uIf{\rm $pyd\_length\_array[i] > 0$ and $ptc\_dist \leq pyd\_length\_array[i]$}{
        \uIf{\rm $ptc.flag$ is 1}{
          $ptc.weight \gets$updateWgt$(ptc, \text{C}_k^\prime\_array, ptcl\_pyd\_array)$ \algorithmiccomment{Update particle weight with (\ref{Eq: particles_posterior_weights1_approx_newborn})}\;
        }
        \uElse{
          $ptc.weight \gets$updateWgtNewBorn$(ptc, \text{C}_k^\prime\_array, ptcl\_pyd\_array)$  \algorithmiccomment{Update newborn particle weight with (\ref{Eq: particles_posterior_weights2_approx_newborn})}\;
        }
      }
  
    }
  }
  \end{algorithm*}
  
  \begin{algorithm*}[t]
  \footnotesize
  \caption{\small Resampling, Occupancy Estimation and  Mixture Model Coefficients Calculation}
  \SetKwProg{Fn}{Function}{}{end}
  \SetKwProg{Struct}{Struct}{}{end}
  \label{Algorithm: Resample and Occupancy Status Estimation}
  
  $occ\_voxel\_array[N_v]$ \algorithmiccomment{This array stores the occupancy probability of each voxel subspace}\;
  
  $dst\_lambda\_array[N_v][2]$ \algorithmiccomment{This array stores the coefficients $\lambda_1$ and $\lambda_2$ of each voxel subspace in DST (Section \ref{Section: Mixture Model})}\;
  
  \For{\rm $i=1:N_v$}{
    $dst\_lambda\_array[i] \gets$ calculateDSTCoefficients$(ptc\_voxel\_array[i])$ \algorithmiccomment{Calculate the DST coefficients with (\ref{Eq: lambda estimation})}\;
    $occ\_voxel\_array[i] \gets$calculateOccPr$(weight\_voxel\_array[i])$ \algorithmiccomment{Calculate a voxel's occupancy probability with (\ref{Eq: occupancy status esimtaion voxel})}\;
  
    $ptc\_voxel\_array[i] \gets$ resample$(ptc\_voxel\_array[i])$ \algorithmiccomment{Resample the particles in a voxel using rejection sampling \cite{RejectionSampling}}\;
  }

  \end{algorithm*}
  
  \begin{algorithm*}[t]
  \footnotesize
  \caption{\small Particle Birth}
  \SetKwProg{Fn}{Function}{}{end}
  \SetKwProg{Struct}{Struct}{}{end}
  \label{Algorithm: New-born}
  
  \For{\rm $i=1:N_f$}{
    \For{\rm $j=1:pyd\_pt\_num\_array[i]$}{
      $pt \gets ptcl\_pyd\_array[i][j]$, $pt\_vel \gets ptcl\_vel\_array[i][j]$ \algorithmiccomment{Traverse every measurement point}\;
      \For{\rm $k=1:L_b$}{
          $pct \gets$ addNoise$(pt)$,  \algorithmiccomment{Add noise to $pt$ according to Section VI-B to generate a new particle}\;
          addParticleToVoxel($pct$), \algorithmiccomment{Store the particle in $ptc\_voxel\_array$}\;
      }
  
    }
  }
  
  \end{algorithm*}
}

\bibliographystyle{IEEEtran}

\bibliography{head}

\end{document}